\documentclass[times,twocolumn,final,authoryear]{elsarticle}

\usepackage{ycviu}
\usepackage{framed,multirow}
\usepackage{subfig}

\usepackage{amssymb}
\usepackage{latexsym}
\usepackage{amsmath}
\usepackage{float}

\usepackage{url}
\usepackage{xcolor}
\definecolor{newcolor}{rgb}{.8,.349,.1}

\journal{Computer Vision and Image Understanding}

\begin{document}

\ifpreprint
  \setcounter{page}{1}
\else
  \setcounter{page}{1}
\fi

\begin{frontmatter}

\title{SnapshotNet: Self-supervised Feature Learning for Point Cloud Data Segmentation Using Minimal Labeled Data }

\author[1]{Xingye \snm{Li}\corref{cor1}} 
\cortext[cor1]{Corresponding author: 
  Tel.: +1-646-719-3732;}
\ead{xli020@citymail.cuny.edu}
\author[1]{Ling \snm{Zhang}}
\author[1,2]{Zhigang \snm{Zhu}}

\address[1]{The City College, The City University of New York, 160 Convent Avenue, New York, NY 10031, USA}
\address[2]{The Graduate Center, The City University of New York, 365 5th Avenue, New York, NY 10016, USA}

\received{1 May 2013}
\finalform{10 May 2013}
\accepted{13 May 2013}
\availableonline{15 May 2013}
\communicated{S. Sarkar}

\begin{abstract}
Manually annotating complex scene point cloud datasets is both costly and error-prone. To reduce the reliance on labeled data, a new model called SnapshotNet is proposed as a self-supervised feature learning approach, which directly works on the unlabeled point cloud data of a complex 3D scene. The SnapshotNet pipeline includes three stages. In the snapshot capturing stage, snapshots, which are defined as local collections of points, are sampled from the point cloud scene. A snapshot could be a view of a local 3D scan directly captured from the real scene, or a virtual view of such from a large 3D point cloud dataset. Snapshots could also be sampled at different sampling rates or fields of view (FOVs), thus multi-FOV snapshots,  to capture scale information from the scene. In the feature learning stage, a new pre-text task called multi-FOV contrasting is proposed to recognize whether two snapshots are from the same object or not, within the same FOV or across different FOVs. Snapshots go through two self-supervised learning steps: the contrastive learning step with both part contrasting and scale contrasting, followed by a snapshot clustering step to extract higher level semantic features. Then a weakly-supervised segmentation stage is implemented by first training a standard SVM classifier on the learned features with a small fraction of labeled snapshots. Then trained SVM is further used to predict labels for input snapshots and predicted labels are converted into point-wise label assignments for semantic segmentation of the entire scene using a voting procedure. The experiments are conducted on the Semantic3D dataset 
and the results have shown that the proposed method is capable of learning effective features from snapshots of complex scene data without any labels. Moreover, the proposed weakly-supervised method has shown advantages when comparing to the state of the art method on weakly-supervised point cloud semantic segmentation.
\end{abstract}


\end{frontmatter}


\section{Introduction}
Studies on 3D point cloud data have been gaining momentum in the field of computer vision.
Deep neural networks such as PointNet\citep{pointnet}, DGCNN\citep{DGCNN} have been proposed for better performances on point cloud related tasks, with the help of larger datasets such as the ModelNet\citep{modelnet} and Semantic3D\citep{semantic3d}.
The collective effort between deep neural networks and dedicated datasets continues to push the state of the art performance on the point cloud object classification.

On the other hand, point cloud semantic segmentation is of great interests in the applications of autonomous driving, robotics and remote sensing\citep{xie2020linking}.
So far most of the deep learning driven point cloud semantic segmentation methods follow the supervised workflow, which requires densely labeled datasets, such as the 1.6 million points Oakland Dataset\citep{oakland}, the 215 million points Stanford Large-scale 3D Indoor Spaces Dataset (S3DIS) dataset\citep{S3DIS} and the 4 billion points Semantic3D\citep{semantic3d}.
However, annotating large scale datasets is at a very high cost both in time and human labors.
This issue is becoming more prominent in applications such as hazard assessment where drive-by and fly-by LiDAR mapping systems have been used to collect massive windstorm damage data sets in recent hurricane events \citep{lidar, terrestrial, gong2013mobile, hu2018advancing}. The fact that LiDAR is starting to be integrated into smaller mobile devices\citep{ipad}, which could lead to a boom in the scale of real life complex point cloud data.

\begin{figure*}[t]
\centering
\includegraphics[scale=0.3]{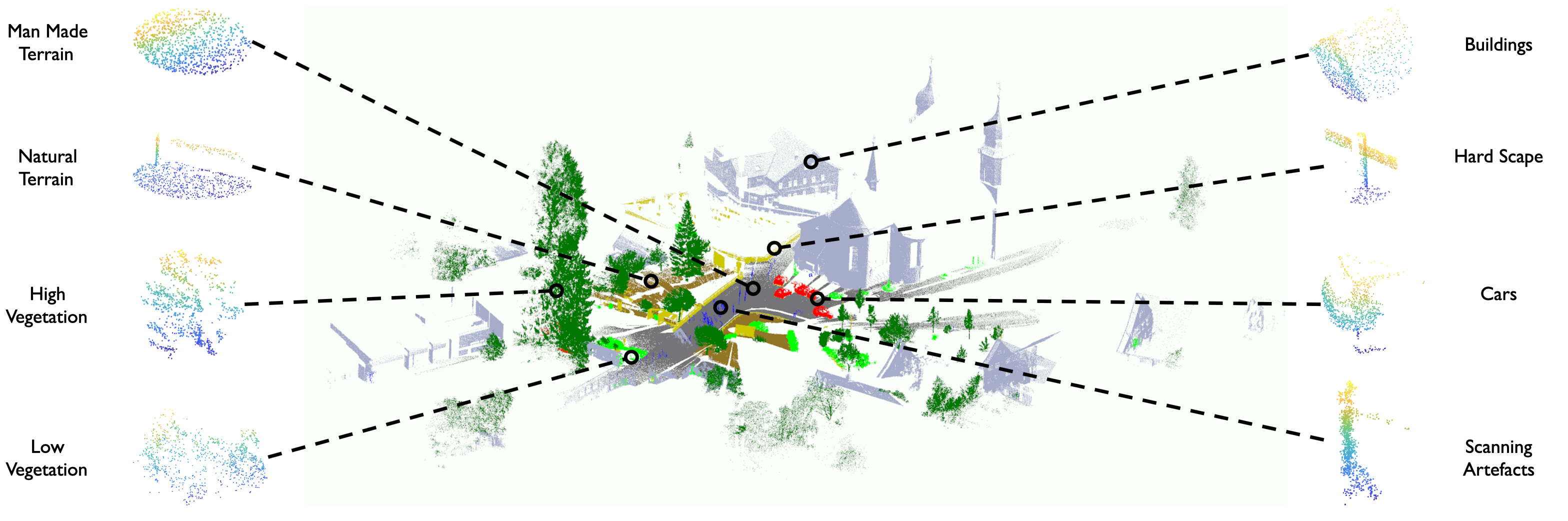}
\caption{Visualization of some snapshots sampled from the Semantic3D dataset. The sampling procedure makes no use of labels, therefore a snapshot may contains points from other classes. The class labels are added manually for visualization}
\label{fig:snapshots}
\end{figure*}

To alleviate the dependence on the labels of large datasets, unsupervised learning methods have drawn increasing attention.
Among the unsupervised methods, one form known as ``self-supervised learning" has been popular in the studies of image data understanding.
This self-supervised approach has found success in designing ``pretext" tasks, such as jigsaw puzzle reassembly\citep{jigsaw}, image clustering\citep{imageClustering} and image rotation prediction\citep{rotation} etc, by training deep learning models for feature extraction without labels being involved.
Based on the idea of solving pretext tasks, we previously developed the model of Contrast-ClusterNet\citep{contrastnet}, which works on unlabeled point cloud datasets by part contrasting and object clustering.
While this work has shown comparable performance to its supervised counterparts on synthetic point cloud objects classification, it inherits the problem of other pretext-driven models used on image data: pretext tasks must be defined regarding the prior knowledge embedded in the data.
In the context of point cloud understanding, the part contrasting and object clustering tasks assume the input data are well separated as individual objects. 
This assumption limits the model's power on real life scene data or where 3D data of single objects cannot be easily obtained.
We addressed this limitation posed on scene point cloud by proposing a snapshot-based method\citep{visapp21}, which captures local clusters of points called snapshots from the scene as input samples to fulfill the tasks of part contrasting and object clustering.
Formally, a snapshot is defined as a collection of points, sampled from a point cloud scene, without knowing their labels or even assuming they are from the the same objects. (Figure \ref{fig:snapshots}). 
It could be a real view of a local 3D scan directly captured from the real scene, or a "virtual" view of such a local 3D scan from a large 3D point cloud dataset.
The effectiveness of this method is evaluated and approved by conducting classification on the captured snapshots with a single FOV in \citep{visapp21}. In this paper, we extend the idea to capture multi-FOV snapshots to further improve the performance of classification.

Besides from the limitation of making assumptions on the training data, another weakness of the Contrast-ClusterNet is that dense labels are still needed for the downstream tasks.
The full supervision involved in the object classification contradicts the main goal of self-supervision, that to save labeling efforts on training data.
To extend the idea of reducing labeled data usage to the downstream tasks, we seek solutions from weak supervision.

Therefore, based on the two pieces of our previous work, we further propose the SnapshotNet, which integrates multi-FOV snapshot generation, contrastive feature learning, and a weakly-supervised technique for point-wise scene segmentation using a voting mechanism. 
First of all, inspired by the observation that, humans are able to distinguish objects at different scales, we present a new pre-text task for contrastive learning, namely multi-FOV contrasting.
When capturing a sample, we take multiple snapshots in different field-of-views (FOVs).
Assuming these multi-FOV snapshots are small enough so they still represent the same object, the task of scale contrasting is to consider whether two snapshots, within one FOV or across multiple FOVs, are of the same object or not. Thus the multi-FOV contrasting includes two parts: part contrasting and scale contrasting.
We will compare the performance of the scale contrasting against the part contrasting, as well as the combination of the two - the multi-FOV contrasting.

For the complete model pipeline, the captured snapshots first go through a two-step self-supervised pipeline using ContrastNet and ClusterNet consecutively for feature learning. 
Then a weakly-supervised approach is implemented by training an SVM classifier on the learned features of a small portion of labeled snapshot samples (mostly cluster centers) combined with the samples automatically labeled from the clusters of the samples generated in the pipeline.
Finally, the entire 3D point cloud scene is repeatedly scanned as random snapshots to go through the feature extractor and classifier. 
The predicted snapshot-wise label is assigned to each point of a snapshot, followed by a voting-based mechanism for the final label for each point.

This work makes the following contributions:

(1) We propose a new contrastive learning method called multi-FOV contrasting, by leveraging point cloud samples at different scales. This task devotes on predicting if two snapshots are of the same object, regardless of their sampling FOVs.

(2) We develop a three-stage approach for semantic segmentation: snapshot generation, self-supervised feature learning, and point-wise segmentation by integrating multiple weakly-supervised classified results.

(3) We study the "purity" of snapshots, and show that the self-supervised learning with impure snapshots can still effectively obtain highly useful semantic features for object classification and scene segmentation. This includes cases when some of the classes do not have well-sampled snapshots.

(4) By using the learned features and clustering to obtain larger pseudo labels with a small number of labels (thus weakly-supervised) to train a simple classifier, we design a simple voting procedure to integrating labels of randomly sample snapshots, which leads to point-wise point cloud scene segmentation performance comparable to the state of the art weakly-supervised methods.

The rest of the paper is organized as the following.
Section \ref{sec:relatedwork} discusses related work on self-supervised learning methods on point cloud, and point cloud semantic segmentation.
Section \ref{sec:method} describes the theory and design of the SnapshotNet for self-supervised feature learning and point cloud semantic segmentation with fewer labeled data.
Section \ref{sec:results} details the experimental results, including the designs and evaluations of data capturing, feature learning, and segmentation. 
Finally Section \ref{sec:conclusion} concludes the work with discussions of a few ideas for future work.

\section{Background and Related Work}
\label{sec:relatedwork}

\subsection{Self-supervised Learning: }
Self-supervised learning aims to predict for output labels that are generated from the intrinsic information of the data.
This topic has been widely studied on the image data where various of pre-text tasks have been proposed, such as context prediction\citep{context_prediction}, jigsaw puzzle reassembly\citep{jigsaw}, image clustering\citep{imageClustering}, and image rotation prediction\citep{rotation} etc, and these methods have demonstrated considerable results on ordered data such as 2D images or videos.

With the advancement in LiDAR technology, the cost for obtaining large scale point cloud data has enormously decreased.
The booming in 3D point cloud data has turned the challenge from data collection to manual annotation, which is much more difficult and laborious compared to 2D data.
To alleviate the use of labeled data, a number of self-supervised models have been proposed lately\citep{achlioptas2018learning, foldingnet, contrastnet, context, xie2020pointcontrast, chen2020unsupervised}.
In previous work of our lab, Zhang and Zhu proposed the Contrast-ClusterNet\citep{contrastnet} with pre-text tasks of first predicting whether two segments are from the same object, leading to the ContrastNet for obtaining self-learned features, which are then used for separating the objects into different clusters using KMeans++, for training another network called ClusterNet to obtain better self-learned features.
The work\citep{contrastnet} has shown the capability of learning features in a self-supervised manner, and then using the features, an SVM classifier can be trained using labeled data for point cloud objects classification.
However, this process still requires to know a set of 3D points belong to a single object (even though the label is not needed). In training the SVMs, the same amount of labeled data as in supervised models is used, therefore decreasing the benefits of leaving out annotations in self-supervised learning.

\subsection{Semantic Segmentation of Point Cloud: }
With the recent works shifting focus to adapting deep learning on LiDAR point cloud data, a series of deep learning based point cloud semantic segmentation methods have been proposed.
As summarized by Guo et al.\citep{survey2020}, there are several mainstream semantic segmentation methods on point cloud data, such as the discretization-based, projection-based, and point-based methods.

The discretization-based approach is greatly inspired by the success of deep learning on 2D grid data, where the 2D data is in a regular representation, in contrast to the unordered 3D point cloud.
A number of works have been proposed using the voxel-based models\citep{rethage2018fully, semantic3d, 3dCNN, segcloud}, which voxelize the point cloud data to 3D grids to enable direct 3D covolutional feature extraction.
Despite that this method has made significant progress on point cloud segmentation, it is very sensitive to the voxel resolution and often has strict requirements on memory and computational power.

The projection-based method, on the other hand, has shown advantages on computation efficiency.
As a representation of this approach, 2D multi-views models are designed to project a 3D point cloud to 2D views from multiple directions, so that traditional convolutional networks can be applied for semantic segmentation tasks\citep{lawin2017deep, boulch2017unstructured, Alonso_2020}.
However, the downside of this approach is that geometrical information is often lost during the dimension reduction.

PointNet\citep{pointnet} is the first deep net proposed to directly work with point cloud data without the pre-processing step of transforming the raw point cloud into voxels or 2D multi-views representations.
To help catching local geometrical context, the PointNet++\citep{pointnet++} is developed by proposing a hierarchical network based on the PointNet.
The idea of exploiting local structures of the 3D data is further explored by developing dynamic graph CNN (DGCNN)\citep{DGCNN}, which uses graphs the geometrical relations of the point cloud and operate convolutions on such graphs.

\begin{figure*}[h]
\centering
\includegraphics[scale=0.2]{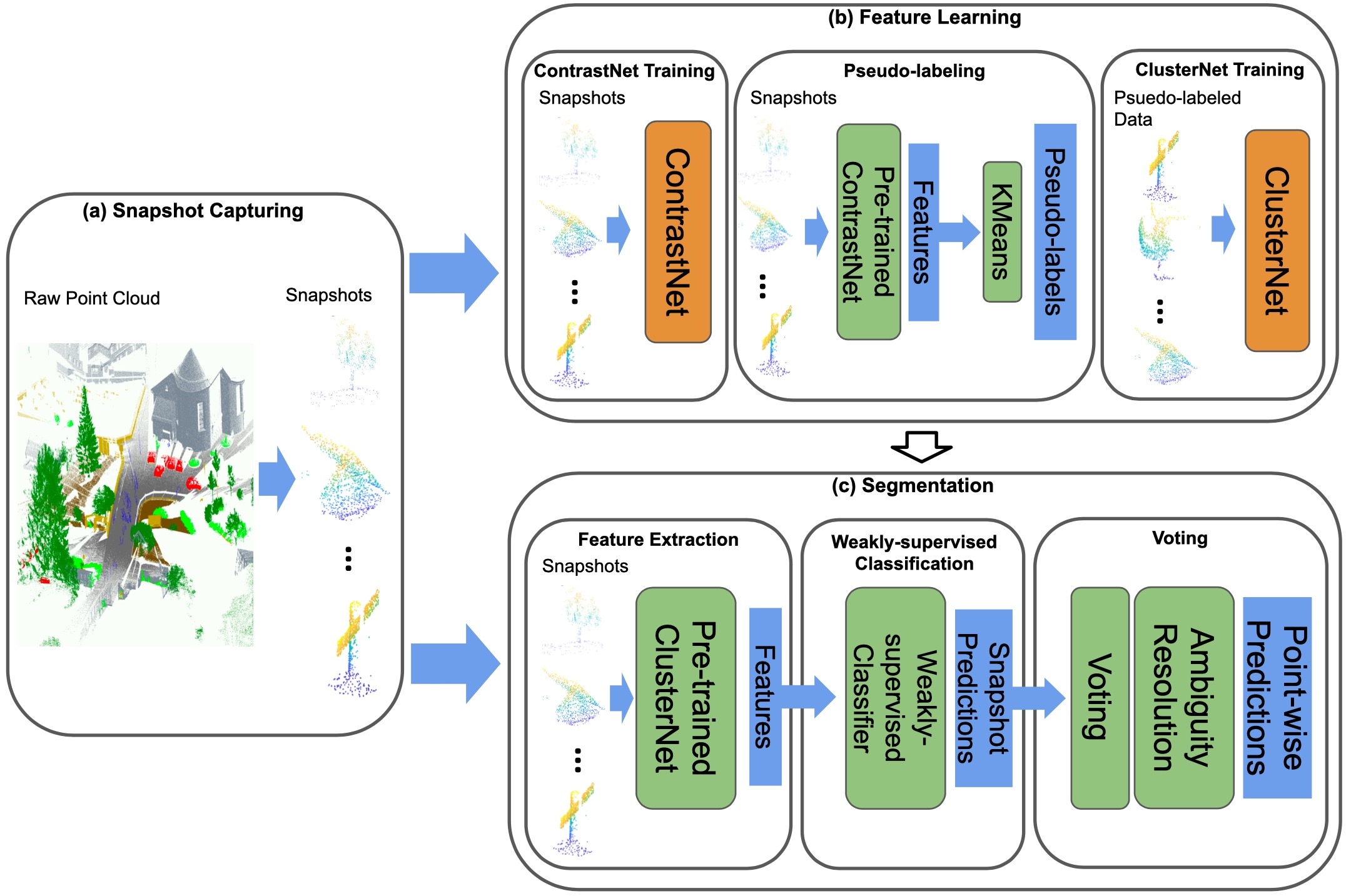}
\caption{The SnapshotNet pipeline: (a) Snapshot capturing from the raw point cloud scenes; (b) Feature learning by conducting contrasting tasks, snapshot clustering and cluster classification; (c) Semantic segmentation by classifying and voting on snapshots.}
\label{fig:pipeline}
\end{figure*}

The above segmentation methods mostly rely on densely labeled data, and such datasets are proven to be costly on time and human labors.
There are few works focusing on weakly-supervised scene point cloud semantic segmentation: 
the segmentation-aided classification\citep{guinard2017weakly} is a non-parametric method, using conditional random field (CRF) to process the output of a pointwise classifier.
The pseudo-labeling approach\citep{pseudo_labelling} trains a PointNet\citep{pointnet} with a handful of labeled points and gradually assigns pseudo-labels that are generated from the trained PointNet model to all unlabeled points, and model is also iteratively updated with more reliable pseduo-lables.
Xu and Lee\citep{weakly} proposed an incomplete supervision model with three additional training losses to constrain the model.
Among them, two pieces of work \citep{guinard2017weakly, pseudo_labelling} that also worked on the outdoor datasets as ours are the baselines that our proposed SnapshotNet will be compared with.

\section{Theory and Design}
\label{sec:method}
Self-supervised learning often requires prior knowledge about the input data to ensure the intrinsic information of the data, from which the labels are derived, is consistent across all samples.
This is also the case of the Contrast-ClusterNet\citep{contrastnet}, which will be used as the base model of our proposed work. As will be summarized below, it has two major modules called ContrastNet and ClusterNet.
Each module is centered on a deep learning neural network DGCNN \citep{DGCNN} capable of extracting features from the point cloud inputs. 
First the ContrastNet takes inputs of paired point cloud segments, which are obtained by randomly cutting the point cloud object into two halves.
The job of the ContrastNet is to consider whether two segments of a pair are from the same object or not, essentially doing the task of binary classification.
The trained ContrastNet is capable of extracting features at high-level due to the nature of the pretext task of part-contrasting.
The second module, the ClusterNet, is to obtain more representative and fine-grained features. 
Before starting training the ClusterNet, features of the raw point cloud objects are extracted by the trained ContrastNet, and these features are subsequently clustered into a much larger number of groups (than the number of object categories) using Kmeans++; in \citep{contrastnet}, experimental studies were also performed for the optimal numbers of clusters.  
Each object is then assigned with their cluster ID as a pseudo-label for the training of the ClusterNet.

Although there are no labels being involved in this two-step feature learning process, the nature of self-supervised learning requires some prior assumptions regarding the pretext tasks that drive the self-supervision.
In this example, such assumption is that each training sample must be an individual point cloud object to enable part contrasting.
This assumption can be easily made on datasets such as ModelNet\citep{modelnet} and ShapeNet\citep{shapenet_data}, where each sample is a synthetic CAD model of a single 3D object.
However, this soon becomes a limitation on real-life point cloud datasets, such as the Okaland\citep{oakland} and Semantic3D\citep{semantic3d} data, where an entire point cloud is a complex scene rather than invidual single objects.

To address this issue, we thus propose the SnapshotNet for the self-supervised feature learning and weakly-supervised semantic segmentation on complex scene point cloud.
As illustrated in Figure \ref{fig:pipeline}, our method consists of three modules: (a)  snapshot capturing, (b) feature learning, and (c) segmentation.
The snapshot capturing procedure, as an analogy to taking snapshots with a 3D camera, captures small areas of the entire point cloud to train the model.
Then the feature learning module uses the Contrast-ClusterNet\citep{contrastnet} as the backbone for self-supervised feature learning.
Finally, in the segmentation module, a classifier is trained on few labeled data and the pseudo-labeled data for snapshot classification.
A voting mechanism is followed to convert snapshot-wise predictions to point-wise predictions, achieving the goal of semantic segmentation.
Each part of the pipeline will be described in details in the following subsections.

\subsection{Snapshot Capturing}

\begin{figure}[H]
\centering
\includegraphics[scale=0.12]{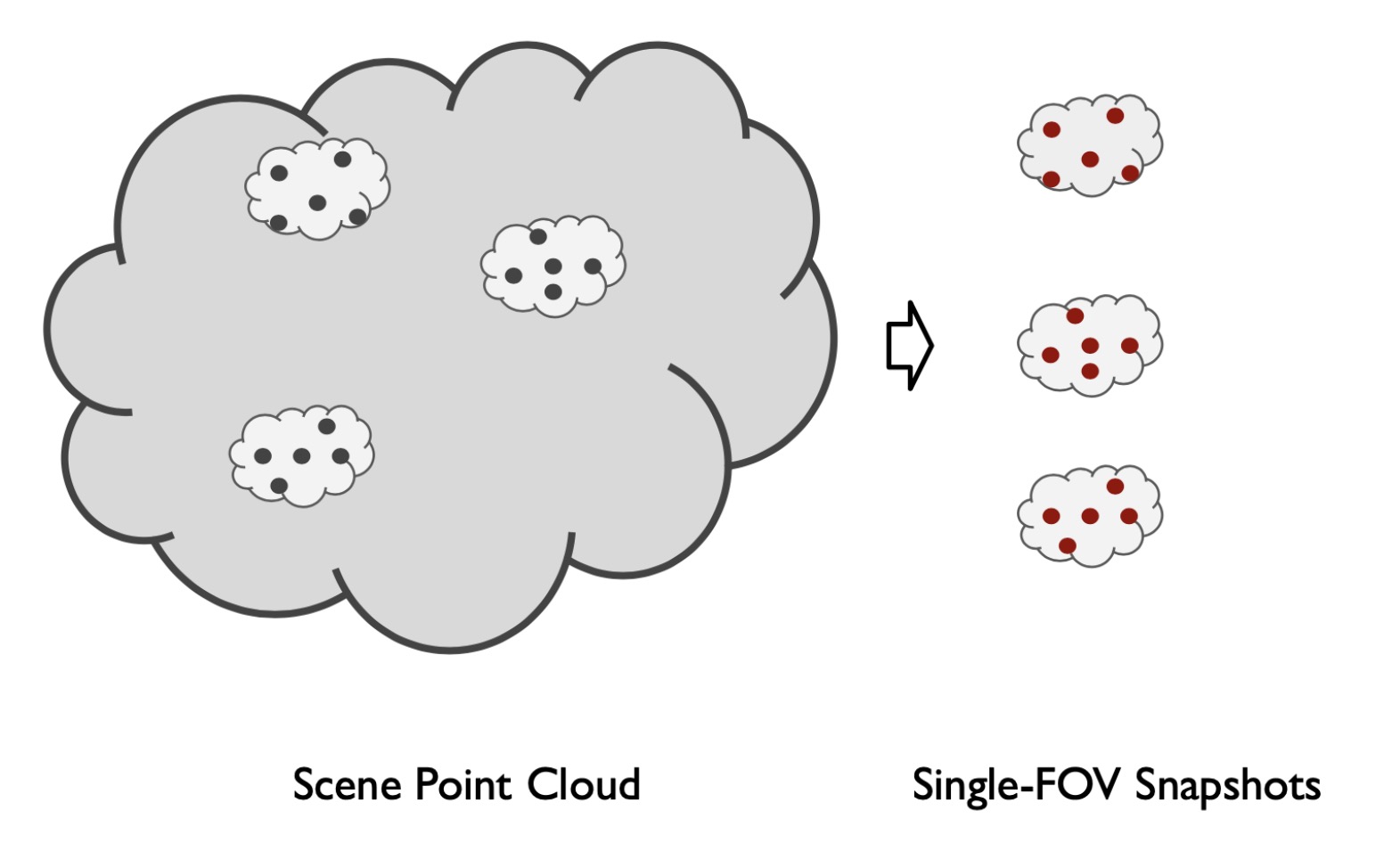}
\caption{Single-FOV snapshots sampling: an illustration.}
\label{fig:singleFOV}
\end{figure}

Given a real-life point cloud dataset, the \textit{snapshot capturing} stage applies random sampling with k-Nearest Neighbors (kNN) to obtain small collections of points as snapshots (Figure \ref{fig:snapshots}).
During each sampling, an anchor point is randomly selected from the point cloud at first, and kNN gives a collection of k points nearest to the anchor point, where k defines the snapshot sampling area. This kNN strategy is a simulation of a virtual snapshot of a local 3D view, followed by point selection based on their 3D proximity to better ensure the sameness of an object.
Each collection is therefore called a `snapshot' of the local neighborhood in the bigger point cloud pool.
\textcolor{black}{Here, thus sampled snapshots share the same sampling rate with the local areas where the anchors are picked up from the scene, meaning that the area covered by each snapshot is determined by the local scene sampling rate.}
In other words, the snapshots have one single field of view (FOV), so they are also notated as the single-FOV snapshots (Figure \ref{fig:singleFOV}).

\textbf{\textit{Purity of snapshots: }}
Since the selection of an anchor point happens randomly in the point cloud, it is possible to have the anchor sitting close to the border between different semantic classes.
This introduces a certain degree of noises to the snapshot by including some points from other minority classes.
Compared to the object based contrasting pretext task, which in this paper is notated as ObjectNet for easy comparison with the SnapshotNet, our method further relaxes the constraint that 3D points of a sample must come from the same object.
The SnapshotNet fundamentally sees each snapshot as a collection of points that represents a small region of the bigger complex scene, where such a collection of points has a high probability of belonging to the same class. 
In our experiment section, we will show how the noises in snapshots will affect the performance of feature learning for later evaluation.

To quantify the noise level of sampled snapshots, we present a metric to evaluate our snapshot sampling quality, namely purity.
When sampling from the Semantic3D\citep{semantic3d}, we utilize the provided labels of the dataset to approximate the semantic label of each snapshot for the sake of snapshot classification evaluation.
A label is assigned to a snapshot by voting from all points that are associated with that sample.
The class label that most points agree on is chosen as the semantic label for the snapshot, the voting procedure can be parameterized as:
\begin{equation}
     C_x = arg\underset{i}{max} \sum_{j=1}^{K} I(y_j = i),
\end{equation}
    
where $x$ is the snapshot sample, $y_j$ is the point-wise label for $x$ ($j$=1,...,$K$), $K$ represents the number of points in the snapshot $x$, and $I$ is an indicator function for the class of each point.
Thus the purity is given by
\begin{equation}
    P(x) = \frac{\sum_{j=1}^{k} I(y_j=C_x)}{k}
\end{equation}
The statistics of the voted semantic labels and the purity for each sample will be further discussed in Section \ref{datasets} using real examples.

\textbf{\textit{Multi-FOV snapshots: }}
Inspired by zooming with a camera while taking a photo, it soon came to us that a different field of view(FOV) of a snapshot image leads to different information content.
Given the same sensor size, a larger FOV might contain more objects at low details, while a smaller FOV focuses on smaller views with greater details.
We adopt this observation into our design of the point cloud snapshots, to include multiple FOVs for each snapshot.

\begin{figure}[H]
\centering
\includegraphics[scale=0.18]{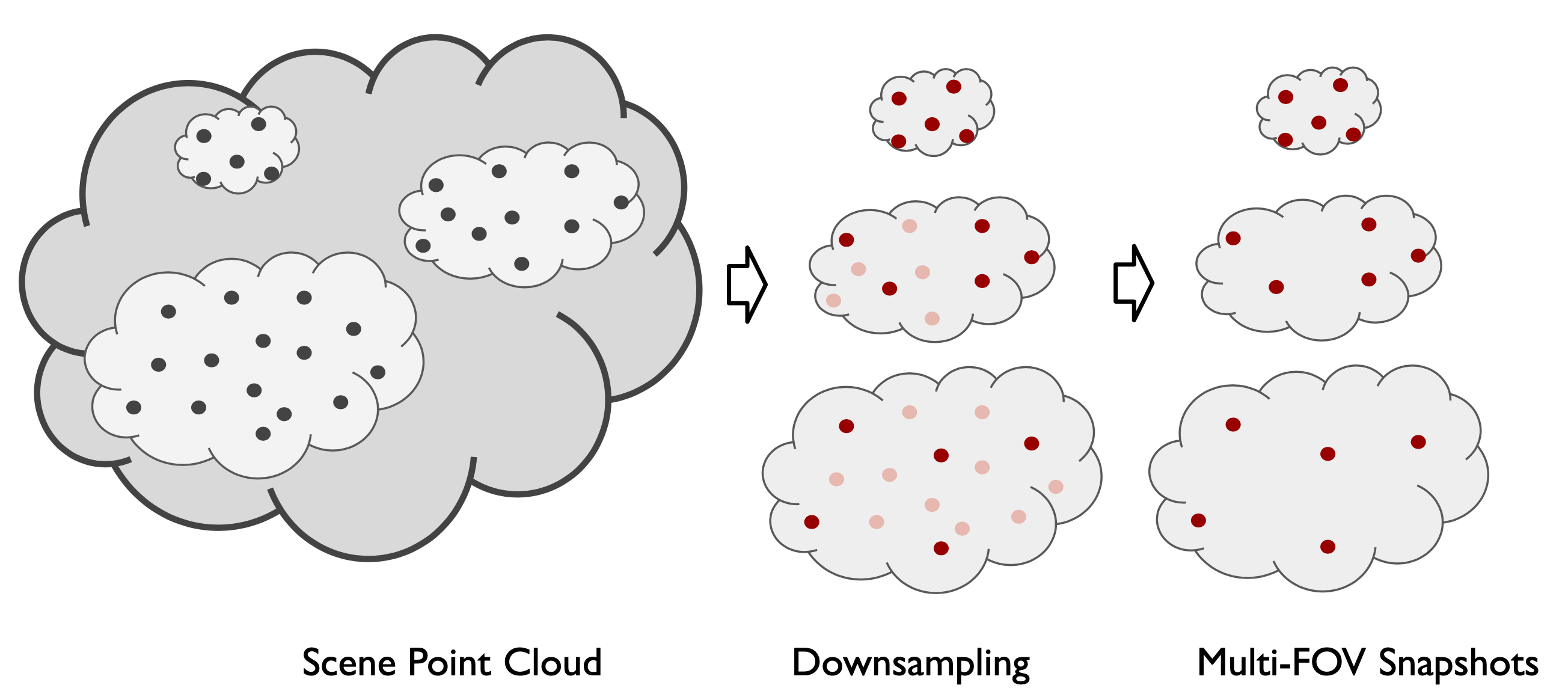}
\caption{Multi-FOV snapshots sampling. Three snapshots of different FOVs are pre-sampled from the scene and each one has a different size. Then the samples are downsampled to meet the same input size of the network, leading to different resolutions.}
\label{fig:multiFOV}
\end{figure}

Just as the sensor size poses limitation on imaging when zooming in or out, the neural network has similar constrains when dealing with samples with different FOVs.
The number of points in each snapshot is kept as the same, and this leaves only the sampling rate to be altered.
Therefore, we keep the original sampling rate of the point cloud as the base, and accordingly decrease the sampling rate by grouping points at a sparser scale.
Specifically, this is achieved by pre-sampling a larger number of points at the base sampling rate and then randomly drop some points to meet the input size (i.e., the number of 3D points).
Figure \ref{fig:multiFOV} illustrates an example of sampling snapshots in three FOVs.
When sampling a snapshot at the base sampling rate, the down-sampling can be well ignored.
However, when capturing a sample with an FOV two times larger than the base FOV, a pre-sampled snapshot of double amount of the points are first captured using kNN.
This pre-sampled snapshot is then downsampled by a factor of two to agree on the network input size while at half the base sampling rate.

There are several intuitions behind this multi-FOV design.
As described earlier, the part-contrasting exploits information at the object level by performing binary classifications on samples that share uniform sampling rate.
On the other hand, the human vision is able to recognize objects at very different scales, which encourages us to further make use of the scale information.
The multi-FOV snapshots are able to fill in the gap of the missing scale information, giving us an edge on contrastive learning by contrasting on various scales in addition to the part-contrasting.
Secondly, the multi-FOV snapshots serves as an approach of data augmentation, to diversify the input data and indirectly making the contrasting learning more challenging to the network.
Furthermore, single sampling rate is inadequate when facing a scene point cloud with objects of various scales and with different sampling resolutions.
This is particularly a problem for the terrestrial scans, where the density of points rapidly changes along the distance to the scanning device.
When the network is trained to take the sampling rate into account, there is the opportunity to explicitly choose an FOV that is more suitable for sampling a specific object from the scene.
For instance, we would want to sample a snapshot of a small object using a small FOV to maximize the purity, and on the other hand to keep a large FOV on larger objects.
This will be discussed in more details in Section \ref{sec:seg}.

\begin{figure*}[!htb]
\centering
\includegraphics[scale=0.28]{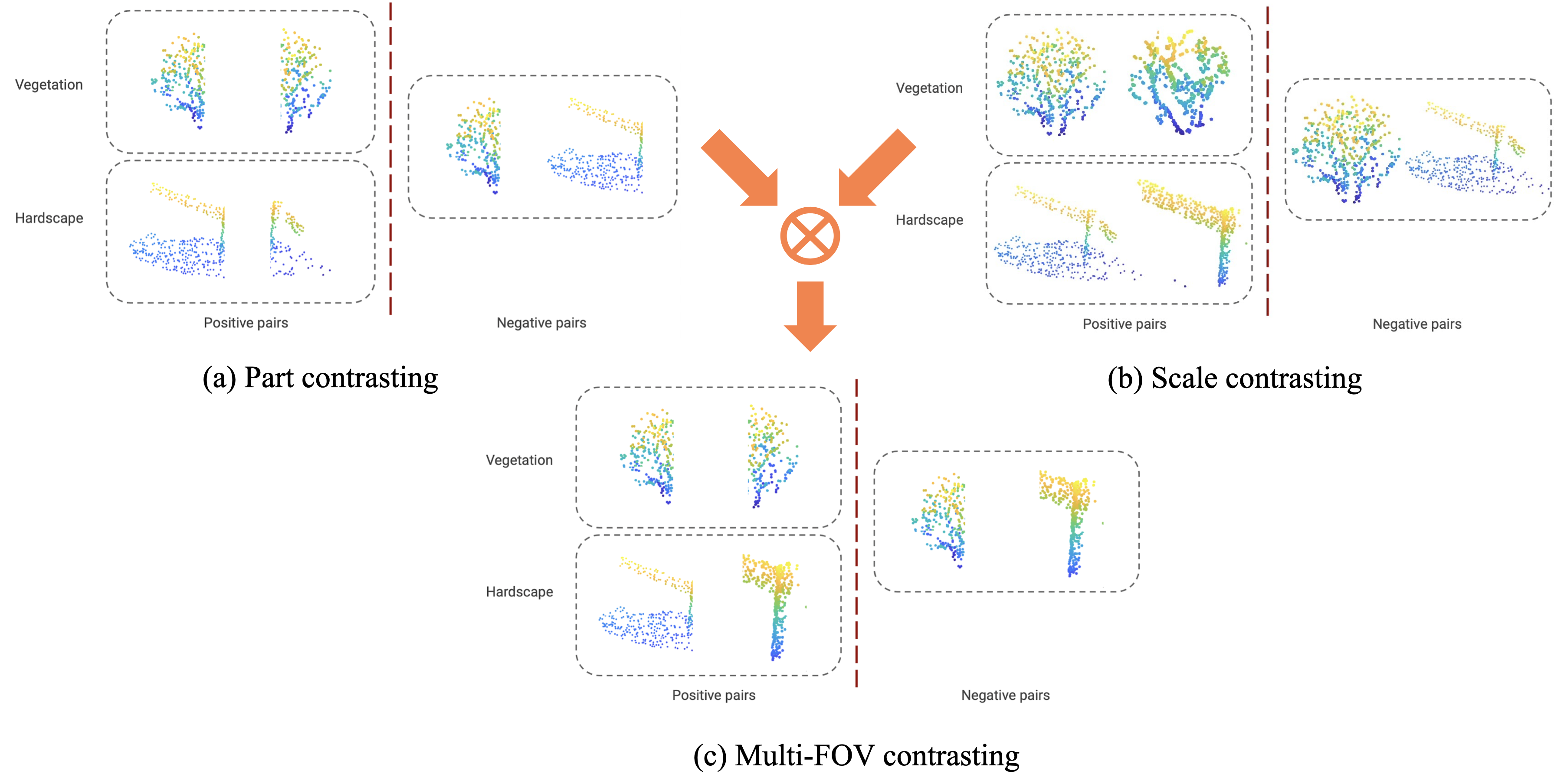}
\caption{Three approaches of contrastive learning by forming positive pairs or negative pairs from two samples. \textcolor{black}{ In the part contrasting (a), each snapshot is split into two parts and then pairs are formed by putting together of any two parts. On the left of (a), the left and right side of the tree form a positive pair because they are from the same snapshot, and the same goes for the fence, whereas on the right of (a) the tree and fence form a negative pair.
The scale contrasting (b) takes two snapshots across different field-of-views as a pair. Here two snapshots of the same tree or same fence (but at different scales) form a positive pair, and a negative pair comes from different objects. The Multi-FOV contrasting (c) combines the previous two approaches by cutting each snapshot into two parts across all FOVs, and forms positive pairs where the two parts are from the same object regardless of their sampling FOVs. A negative pair has two parts from different objects.}}
\label{fig:contrastive_learning}
\end{figure*}

\subsection{Self-supervised Learning with Snapshots}
After being captured from the scene, each `snapshot' is viewed as a single point-cloud object and fed into the two-stage ContrastNet-ClusterNet for \textit{feature learning}. 
Both networks are based on the DGCNN\citep{DGCNN}, therefore they are similar to each other in structures.

\textbf{\textit{Contrastive feature learning:}}
The contrastive learning includes three approaches: part-contrasting, scale-contrasting, and multi-FOV contrasting.
When conducting the \textit{part-contrasting} during the training of a ContrastNet, we follow the random cutting procedure as described in \citep{contrastnet}: two segments from the same snapshot make up a positive pair, which is labeled as 1, and on the contrary, and a negative pair consists of two segments from two different snapshots is labeled as 0 (Figure \ref{fig:contrastive_learning} (a)).
The ContrastNet then learns to recognize whether the input pair is positive or negative, and the parameters are optimized by the Adam optimizer on the cross-entropy loss.

The part-contrasting considers the similarity between different parts of an object in its single-FOV snapshot, thus learning fine-grained features.
The \textit{scale-contrasting}, on the other hand, attempt to learn higher-level features for representing similarity between snapshots of an object across different scales (i.e., with different FOVs).
For instance, the details of an object might get lost in a very small FOV, yet the model is still required to correctly connect this sample to its large FOV counterparts without these details.
To implement this method, we similarly make up pairs from the multi-FOV snapshots: two snapshots in whichever FOVs sampled from the same anchor point form a positive pair, given a label as 1. 
Two snapshots from two different sampling anchors form a negative pair with a label of 0, as shown in figure \ref{fig:contrastive_learning} (b).

The part contrasting and scale contrasting focus on very dissimilar goals, but leading to different levels of features.
However, these two pre-texts are not mutually exclusive when governing the self-supervised learning.
Our design of the multi-FOV snapshots provides additional room to join these two tasks when forming the training sample pairs, and we name this combination as \textit{multi-FOV contrasting}.
Now a positive pair is not limited to coming from two segments of the same single-FOV sample, we can also take two cross-FOV segments from the same sample as a positive pair, and vice versa for a negative pair (Figure \ref{fig:contrastive_learning} (c)).
This formation of sample pairs is expected to push the model into learning both fine-grained and high-level abstract features.

\textbf{\textit{Clustering for feature refinement:}}
Once the ContrastNet is well trained with one of these pre-texts, we continue to adopt the idea of knowledge transfer for more refined features by learning similarities and differences of samples across different snapshots.
Before starting training the ClusterNet, the learned features from the ContrastNet are used to cluster the snapshot samples into k groups with KMeans++. These cluster (group) labels are treated as pseudo-labels for the snapshots to train the ClusterNet.
We use k=300 to cluster the snapshots of all FOVs, into new classes based on the studies in our previous work \citep{contrastnet}.
Note that this number is much greater the number of the existing semantic labels in the Semantic3D dataset (which is eight); however the large cluster number forces the ClusterNet to learn fine-grained features. 
The loss function defined in the work of the ClusterNet\citep{contrastnet} is described as:
\begin{equation}
    \underset{\theta , W}{min}\frac{1}{N}\sum_{n=1}^{N}l (g_{W}(f_{\theta }(x_{n})),y_{n})
\end{equation}

\noindent{where the $g_{W}$ is the classifier that predicts for the correct pseudo-labels $y_n$ given the features $f_{\theta }(x_{n})$.}

\subsection{Semantic Segmentation with Snapshots}
The semantic segmentation has three major components, namely feature extraction, weakly-supervised classification, and point-wise semantic segmentation via voting ((c) in figure \ref{fig:pipeline}).
The feature extraction step is a straightforward process that takes snapshots captured from the raw point cloud and extracts the deep features using the already trained ClusterNet, as described above.
We then use a small fraction of the extracted features along with their labels to train an SVM classifier.
This classifier serves two purposes: one is to evaluate the self-supervised features learned by the SnapshotNet in the experiments, and the second purpose is to serve as a base classifier that will further diffuse all snapshot predictions into point-wise predictions.

\textbf{\textit{Classification with weak supervision and pseudo labeling:}}
Following the self-supervised feature learning, a classifier is trained on the extracted features of labeled training data for classification.
Conventionally, the training process requires as many labeled data as possible for better performance.
However, this approach in its essence is in contradiction with the objective of self-supervised learning, which aims to reduce the dependency on labeled data.
Therefore, we seek solutions from weak supervision to reduce the reliance on dense labels for the downstream tasks following our self-supervised feature learning.

Here the weak supervision can be viewed from two perspectives.
First is that when there are only a few labels available, we still wish to achieve comparable classification performance with the limited labels.
furthermore, the labels assigned to the snapshots are essentially coarse-grained labels because instead of point-wise labeling, each snapshot is labeled as a whole, regardless of the noises included during the sampling.

\begin{figure*}[t!]%
\subfloat[Clusters vs. Semantic Labels]{
\begin{minipage}[c]{0.48\textwidth}
   \centering
   \includegraphics[width=0.98\textwidth]{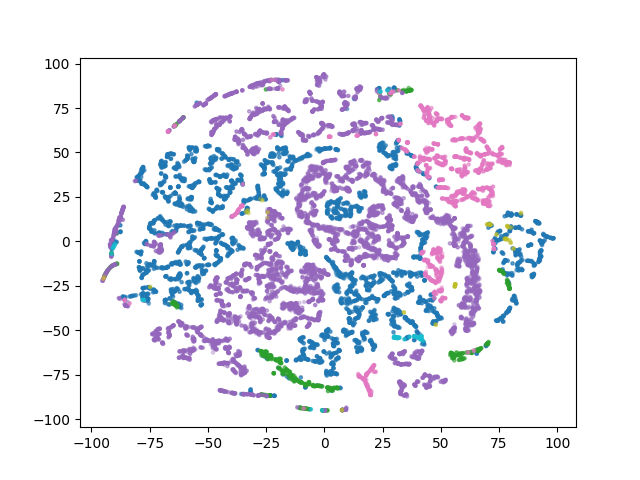}
\end{minipage}}
\subfloat[Clusters vs. Pseudo Labels]{
\begin{minipage}[c]{0.48\textwidth}
   \centering
   \includegraphics[width=0.98\textwidth]{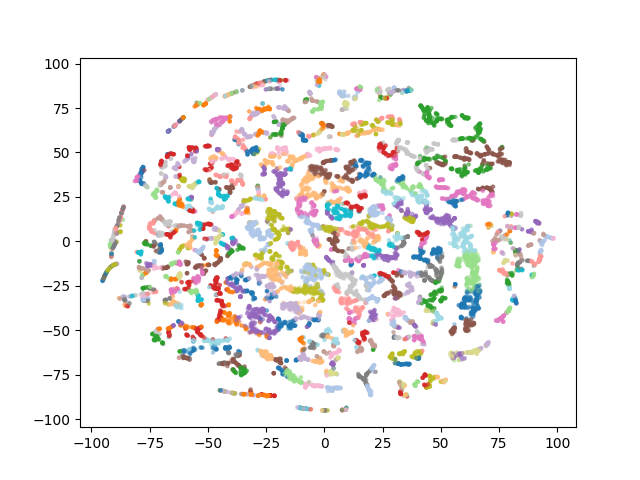}
\end{minipage}}
\caption{Visualization of the feature embedding of the clustered snapshots from the Semantic3D. The clusters are colored by the semantic labels in 6 classes in (a), and by the pseudo labels in 300 classes in (b).\textcolor{black}{ The semantic labels in (a) are represented by the following colors: terrain - blue, vgetaion - green, building - purple, hard scape - pink, artefacts - yellow, and cars - cyan.} It is shown that samples sharing the same pseudo labels are likely to have the same semantic labels as well. \textcolor{black}{Due to the large number of clusters in (b), it is not possible to distinguish all 300 colors globally, but it is possible to distinguish local clusters by their color contrasts. As a support of our claim, the average percentage of points agree with their semantic labels in each cluster among all 300 clusters is 91.67\%.} }
\label{clusters}
\end{figure*}

Second, a pseudo labeling technique is proposed to acquire larger training data population to feed the classifier. This technique is incorporated into the KMeans++ clustering in the feature learning module, hence named cluster-based pseudo labeling. 
Figure \ref{clusters} visualizes the 300 clusters of training samples using KMeans++ in the feature space against their pseudo labels. 
Due to the large number of clusters, each one of them facilitates only a few to a few hundreds of samples.
This large collection of clusters breaks all samples into smaller groups by their similarities in the feature space, where each group hosts way less but highly alike samples. 
This can be seen from Figure \ref{clusters}, when visualizing the clusters against their semantic labels.
This property can be well used by just giving one label to each cluster and assign this label as the pseudo labels of some of the most related samples to that labeled sample in the cluster center.
The selection of the nearby samples can be designed to work geometrically or statistically.
In this work, a threshold is introduced to constrain the measurement of the normalized distance between each point to their cluster center.
A strict threshold filters out samples far from the center to gain more accurate pseudo labeling.

\textbf{\textit{Semantic segmentation by voting:}}
Associated with our first assumption that, statistically a snapshot is able to represent a small piece of an object, we can assume that all points included in a snapshot are highly likely to belong to the same class predicted by the classifier for this snapshot.
The predicted class label is assigned all points in the snapshot.
Thus the point-wise segmentation problem is converted into an object classification problem.
Statistically, if the snapshot capturing happens randomly, all snapshots are able to cover the whole scene after certain number of iterations.
Therefore by repeating the capture-predict-assign procedure, all points in a scene eventually get a predicted label.

The model keeps count of the points with at least one prediction to track the progress of segmentation, and a cut off threshold is set to stop the snapshot capturing.
When $99.95\%$ of the points are assigned with a prediction, the model stops taking new samples and moves to the next step of voting.
Due to the randomness of the snapshot capturing, it is expected to have multiple snapshots covering the same points from the scene, which potentially assigns multiple labels to one point.
To reach for a final agreement on the label, a voting procedure is designed to select the dominant label with most counts to be the final decision for each point.
It is also possible that some points obtained equal numbers of different labels by the time the snapshot capturing stops.
For points with such labeling conditions and particularly near the boundary of different semantic classes, it is likely two labels have the most counts at the same time and the voting would turn into a 50\% chance dice rolling.
To solve this problem, before the final voting we search through the whole point cloud and collect one more label by voting through kNN (k=5) for those with even number of votes.

\textbf{\textit{Multi-FOV snapshots for speed and accuracy:}}
\label{sec:seg}
So far the Multi-FOV snapshots have been participating in the network training.
Yet another important role of the multi-FOV design is to enable faster segmentation and more precise snapshot sampling leveraging our adaptive sampling technique.
The adaptive sampling works to choose one of the pre-listed FOVs according to the size of objects being sampled.
This process is completed in three steps: variance estimation, FOV inquiry, and snapshot down-sampling.

\textit{Variance estimation} refers to the procedure of measuring how spread out the associated points are in a snapshot sample: we take the mean from each of the X, Y, Z coordinates as an imaginary center point for one pre-sampled snapshot and compute the sampling variance from the center point.
Note that we use the largest FOV from the list to pre-sample a snapshot and keep the corresponding variance, with the intention to adequately differentiate the variances by maximizing the sampled area.
Next is to \textit{inquire} the most appropriate FOV for each pre-sampled snapshot based on the variance and the sampling history.
During the segmentation progress, variances from all pre-sampled snapshots are kept to periodically update a KMeans for clustering, where k equals to the number of the pre-listed FOVs.
Before the KMeans is sufficiently trained, the model selects the smallest FOV for the next step of down-sampling.
The reason for this is that the KMeans at this stage does not own enough history records to make a meaningful decision on which FOV to utilize, so the model proceeds with the most conservative option (a small FOV) to ensure a less risky and noisy down-sampling.
Once the KMeans is well optimized, the model starts to inquire for cluster ID by sending in the pre-sample variance, and each cluster ID represents one of the FOVs that will be used for final sampling.
In the end, the \textit{down-sampling} follows the same principle of multi-FOV snapshots sampling, where a larger snapshot is first obtained with the assigned FOV using kNN, before points are randomly discarded to meet the network input size.

It is common to have different outdoor objects at a great range of scales.
When sampling from a large regular surface, such as the ground or the building facade, the chances of including points from other objects is smaller.
This observation motivates us to exploit the advantages of the adaptive sampling for a faster segmentation process.
To allow this, both the down-sampled snapshot for label prediction and the pre-sampled snapshot for segmentation are kept.
It can be seen from Figure \ref{fig:multiFOV} that, the pre-sampled snapshots cover the same area as their corresponding down-sampled snapshots but include more points, except for the smallest FOV.
Here, during the segmentation, when a prediction is acquired from the SVM, the model assigns the prediction to all points in the pre-sampled snapshot instead of the down-sampled one.

This serves as a solution to the low efficiency caused by the down-sampling operation.
The discarded points are highly likely to come from the same class as their neighbor points obtained a prediction, but they won't be given a label until next time they are pre-sampled again and survived the down-sampling to go through the network, which is redundant as repetitive operations.
Now that it is possible to expedite the segmentation of a large uniformed surface at a lower cost, we can leverage even smaller snapshot size to capture local structures at a higher precision.
Overall, the adaptive sampling helps increasing the sampling precision for small objects while maintaining a fast segmentation speed for large surfaces.

\section{Experiments and Results}
\label{sec:results}
Extensive experiments are conducted to evaluate the effectiveness of our proposed approach for both self-supervised feature learning and weakly-supervised point cloud semantic segmentation.
The implementation and experimental results are described in details in the following sub-sections, \textcolor{black}{including (1) datasets used in our experiments; (2) variations of snapshot capturing and their evaluations; (3) self-supervised feature learning by point cloud classification with fewer labeled data; and (4) evaluation of semantic segmentation by point-wise voting.}

\subsection{\textcolor{black}{Datasets}}
\label{datasets}

\begin{table*}[t]
\centering
\footnotesize
\title{Statistics of Snapshot Sampling on Semantic3D dataset}
\begin{tabular}{l|c|ccccccccc}
\hline
 &
  Metrics &
  \begin{tabular}[c]{@{}c@{}}Man Made \\ Terrain\end{tabular} &
  \begin{tabular}[c]{@{}c@{}}Natural\\ Terrain\end{tabular} &
  \begin{tabular}[c]{@{}c@{}}High \\ Vegetation\end{tabular} &
  \begin{tabular}[c]{@{}c@{}}Low \\ Vegetation\end{tabular} &
  Buildings &
  \begin{tabular}[c]{@{}c@{}}Hard\\ Scape\end{tabular} &
  \begin{tabular}[c]{@{}c@{}}Scanning\\ Artefacts\end{tabular} &
  Cars &
  Total \\ \hline
\multirow{2}{*}{Scene 1} &
  Purity(\%) &
  \begin{tabular}[c]{@{}c@{}}98.42\\ $\pm$0.27\end{tabular} &
  \begin{tabular}[c]{@{}c@{}}59.42\\ $\pm$6.35\end{tabular} &
  \begin{tabular}[c]{@{}c@{}}96.15\\ $\pm$2.51\end{tabular} &
  \begin{tabular}[c]{@{}c@{}}92.57\\ $\pm$2.05\end{tabular} &
  \begin{tabular}[c]{@{}c@{}}99.47\\ $\pm$0.15\end{tabular} &
  \begin{tabular}[c]{@{}c@{}}97.02\\ $\pm$0.77\end{tabular} &
  \begin{tabular}[c]{@{}c@{}}78.01\\ $\pm$29.38\end{tabular} &
  \begin{tabular}[c]{@{}c@{}}87.31\\ $\pm$5.23\end{tabular} &
  \begin{tabular}[c]{@{}c@{}}98.37\\ $\pm$0.19\end{tabular} \\ \cline{2-11} 
 &
  Samples &
  \begin{tabular}[c]{@{}c@{}}2891.91\\ $\pm$92.17\end{tabular} &
  \begin{tabular}[c]{@{}c@{}}21.03\\ $\pm$9.57\end{tabular} &
  \begin{tabular}[c]{@{}c@{}}83.25\\ $\pm$21.08\end{tabular} &
  \begin{tabular}[c]{@{}c@{}}256.3\\ $\pm$35.54\end{tabular} &
  \begin{tabular}[c]{@{}c@{}}3847.68\\ $\pm$92.13\end{tabular} &
  \begin{tabular}[c]{@{}c@{}}838.01\\ $\pm$61.28\end{tabular} &
  \begin{tabular}[c]{@{}c@{}}10.34\\ $\pm$6.80\end{tabular} &
  \begin{tabular}[c]{@{}c@{}}51.48\\ $\pm$16.30\end{tabular} &
  8000 \\ \hline
\multirow{2}{*}{Scene 2} &
  Purity(\%) &
  \begin{tabular}[c]{@{}c@{}}94.39\\ $\pm$0.81\end{tabular} &
  \begin{tabular}[c]{@{}c@{}}96.42\\ $\pm$0.37\end{tabular} &
  \begin{tabular}[c]{@{}c@{}}95.03\\ $\pm$0.76\end{tabular} &
  \begin{tabular}[c]{@{}c@{}}91.59\\ $\pm$1.23\end{tabular} &
  \begin{tabular}[c]{@{}c@{}}94.22\\ $\pm$1.10\end{tabular} &
  \begin{tabular}[c]{@{}c@{}}94.40\\ $\pm$0.78\end{tabular} &
  - &
  \begin{tabular}[c]{@{}c@{}}85.58\\ $\pm$2.93\end{tabular} &
  \begin{tabular}[c]{@{}c@{}}94.96\\ $\pm$0.29\end{tabular} \\ \cline{2-11} 
 &
  Samples &
  \begin{tabular}[c]{@{}c@{}}1139.28\\ $\pm$63.40\end{tabular} &
  \begin{tabular}[c]{@{}c@{}}3264.49\\ $\pm$81.99\end{tabular} &
  \begin{tabular}[c]{@{}c@{}}1075.44\\ $\pm$71.23\end{tabular} &
  \begin{tabular}[c]{@{}c@{}}593.05\\ $\pm$51.52\end{tabular} &
  \begin{tabular}[c]{@{}c@{}}569.86\\ $\pm$51.23\end{tabular} &
  \begin{tabular}[c]{@{}c@{}}1240.92\\ $\pm$76.18\end{tabular} &
  0 &
  \begin{tabular}[c]{@{}c@{}}116.96\\ $\pm$28.20\end{tabular} &
  8000 \\ \hline
\end{tabular}
\caption{Statistics of single-FOV snapshot sampling on the two scenes from the Semantic3D dataset (Scene 1:  `Untermaederbrunnen3'; Scene 2:`Bildstein3'). The sampling  is conducted 100 times, with 8000 samples per time. `-' indicates no snapshots being sampled.}
\label{semantic3d_stats}
\end{table*}

\begin{table*}[t]
\centering
\footnotesize
\begin{tabular}{cccccccccc}
\hline
\multicolumn{1}{c|}{\multirow{2}{*}{Method}} &
  \multicolumn{1}{l|}{\multirow{2}{*}{Overall Accuracy (\%)}} &
  \multicolumn{8}{c}{Per-class Accuracy (\%)} \\ \cline{3-10} 
\multicolumn{1}{c|}{} &
  \multicolumn{1}{l|}{} &
  \begin{tabular}[c]{@{}c@{}}Man Made \\ Terrain\end{tabular} &
  \begin{tabular}[c]{@{}c@{}}Natural\\ Terrain\end{tabular} &
  \begin{tabular}[c]{@{}c@{}}High \\ Vegetation\end{tabular} &
  \begin{tabular}[c]{@{}c@{}}Low \\ Vegetation\end{tabular} &
  Buildings &
  \begin{tabular}[c]{@{}c@{}}Hard\\ Scape\end{tabular} &
  \begin{tabular}[c]{@{}c@{}}Scanning\\ Artefacts\end{tabular} &
  Cars \\ \hline
\multicolumn{10}{c}{100\% training data} \\ \hline
\multicolumn{1}{c|}{DGCNN} &
  \multicolumn{1}{c|}{86.5} &
  98.41 &
  - &
  100 &
  85.71 &
  84.38 &
  47.14 &
  - &
  20 \\
\multicolumn{1}{c|}{ObjectNet} &
  \multicolumn{1}{c|}{97.88} &
  97 &
  98 &
  100 &
  92 &
  98 &
  \textbf{98} &
  100 &
  \textbf{100} \\
\multicolumn{1}{c|}{\begin{tabular}[c]{@{}c@{}}SnapshotNet on objects\end{tabular}} &
  \multicolumn{1}{c|}{\textbf{98.63}} &
  \textbf{100} &
  \textbf{99} &
  100 &
  92 &
  \textbf{100} &
  \textbf{98} &
  100 &
  \textbf{100} \\
\multicolumn{1}{c|}{\begin{tabular}[c]{@{}c@{}}SnapshotNet on snapshots\end{tabular}} &
  \multicolumn{1}{c|}{97.5} &
  99.68 &
  - &
  100 &
  \textbf{94.29} &
  98.63 &
  84.29 &
  - &
  80 \\ \hline
\multicolumn{10}{c}{20\% training data} \\ \hline
\multicolumn{1}{c|}{DGCNN} &
  \multicolumn{1}{c|}{84.13} &
  \textbf{99.68} &
  - &
  0 &
  85.71 &
  79.45 &
  55.71 &
  - &
  0 \\
\multicolumn{1}{c|}{ObjectNet} &
  \multicolumn{1}{c|}{94.88} &
  92 &
  \textbf{99} &
  99 &
  85 &
  95 &
  \textbf{95} &
  95 &
  \textbf{99} \\
\multicolumn{1}{c|}{\begin{tabular}[c]{@{}c@{}}SnapshotNet on objects\end{tabular}} &
  \multicolumn{1}{c|}{95.63} &
  98 &
  95 &
  \textbf{100} &
  \textbf{89} &
  91 &
  \textbf{95} &
  \textbf{100} &
  97 \\
\multicolumn{1}{c|}{\begin{tabular}[c]{@{}c@{}}SnapshotNet on snapshots\end{tabular}} &
  \multicolumn{1}{c|}{\textbf{97.13}} &
  99.37 &
  - &
  90 &
  85.71 &
  \textbf{99.45} &
  85.71 &
  - &
  40 \\ \hline
\multicolumn{10}{c}{5\% training data} \\ \hline
\multicolumn{1}{c|}{DGCNN} &
  \multicolumn{1}{c|}{72.88} &
  54.33 &
  - &
  0 &
  62.86 &
  89.86 &
  \textbf{92.86} &
  - &
  0 \\
\multicolumn{1}{c|}{ObjectNet} &
  \multicolumn{1}{c|}{88.38} &
  82 &
  \textbf{93} &
  98 &
  71 &
  82 &
  89 &
  \textbf{95} &
  \textbf{97} \\
\multicolumn{1}{c|}{\begin{tabular}[c]{@{}c@{}}SnapshotNet on objects\end{tabular}} &
  \multicolumn{1}{c|}{90.13} &
  90 &
  87 &
  \textbf{100} &
  \textbf{84} &
  86 &
  87 &
  92 &
  95 \\
\multicolumn{1}{c|}{\begin{tabular}[c]{@{}c@{}}SnapshotNet on snapshots\end{tabular}} &
  \multicolumn{1}{c|}{\textbf{95.0}} &
  \textbf{97.46} &
  - &
  \textbf{100} &
  68.57 &
  \textbf{98.63} &
  78.57 &
  - &
  80 \\ \hline
\end{tabular}

\caption{Classification performance on the snapshots and labeled objects from the Semantic3D dataset, using the DGCNN, ObjectNet and SnapshotNet. Note that `-' means no samples from that class are obtained for testing.}
\label{scene1}
\end{table*}

\textcolor{black}{The experiments are mostly conducted on the Semantic3D large scale point cloud classification benchmark\citep{semantic3d}.}
This dataset consists of a variety of scenes across eight classes: man-made terrain, natural terrain, high vegetation, low vegetation, buildings, hard scape, scanning artefacts, and cars (Figure \ref{fig:snapshots}).
\textcolor{black}{Considering the huge scale and high density} of this dataset is beyond our computational capacity, we choose two scenes, named `untermaederbrunnen3' and `bildstein3' for our experiments, which consist of 27.9 million points and 7.9 million points, respectively.
We use all eight labels to conduct snapshot classification for the evaluation of the self-supervised feature learning.
To evaluate our weakly-supervised semantic segmentation on this dataset, we follow the experiment settings from the state of the art methods\citep{guinard2017weakly, pseudo_labelling}: to combine the man made terrain and the natural terrain into a single class of terrain, and merge the high vegetation and low vegetation into vegetation.

\textcolor{black}{Another dataset incorporated in the semantic segmentation experiments is the Oakland dataset\citep{oakland}. 
In comparison to Semantic3D, the Oakland data has significantly lower but more uniform point densities as scanned from a moving platform.
It has five semantic classes: scatter misc, default wire, utility pole, load bearing and facade.
The class labels are drastically unbalanced, and out of the total 1.4 million points only 7000 points are default wire and 10000 points are utility poles, while the load bearing has near 1 million points.}

\subsection{\textcolor{black}{Snapshots Capturing}}
\textbf{\textit{Snapshot generation:}}
In this experiment, for evaluating the self-supervised feature learning, we capture 8000 single-FOV snapshots \textcolor{black}{from the Semantic3D dataset} as training set and 800 samples for testing, with 1024 points in each sample.
In an ideal setup where the dataset is perfectly balanced between classes, we would obtain close to 1000 training snapshots each class, but this can hardly be realized in real world scenarios (see Table \ref{semantic3d_stats}, \# of samples).
The high resolution of the Semantic3D dataset \citep{semantic3d} poses a dilemma during the single-FOV snapshot sampling: a small sampling size is insufficient to capture details while a large sample brings burdens to the computations.
A compromise is made here, which takes a similar approach to the multi-FOV snapshots sampling, to take a pre-sampled snapshot with 10 times of the network input size (10240 points) and down-sample back to 1024 points per snapshot.

For the semantic segmentation, the same amount of multi-FOV snapshots are captured from both datasets at a smaller size of 512 points as the training data.
The snapshot generation follows the multi-FOV sampling as described previously (Figure \ref{fig:multiFOV}), and we choose three FOVs for each snapshot.
The original sampling rate is chosen as the base sampling rate, and the other two larger FOVs are respectively two times and ten times to the base sampling rate.
Note that since each multi-FOV sample has three FOVs, the total training samples are 24000.

\textbf{\textit{Snapshot purity:}}
Based on the proposed purity metric, we run the single-FOV snapshot capturing procedure on the Semantic3D at the base sampling rate 100 times for statistics.
As shown in Table \ref{semantic3d_stats}, the snapshot purity of a class is correlated with the numbers of sampled snapshots.
From the point-wise perspective, when choosing an anchor point to find the nearest neighbors, each point in the scene has equal chance being selected as the anchor.
However, class-wise speaking, when collecting points surrounding an anchor from a smaller class (i.e., a class with smaller number of points in the scene), the chance of including inter-class points is relatively higher than for a larger class, and this potentially leads to lower purity on smaller classes.
Despite the fact that noises are much more likely to be included in smaller class samples, we can still see that the overall snapshot purity is above $90\%$.
This result is in favor of our claim that, statistically each snapshot is highly capable of representing a small piece of one class from the whole point cloud. 
Nevertheless, we will also investigate if low purity classes can also be fairly treated.

\textbf{\textit{Snapshots versus objects:}}
To comparatively evaluate our self-supervised feature learning, we also apply the same single-FOV sampling procedure on points grouped by the original semantic labels, instead of the whole scene of point cloud.
As a result, these samples obtained exclusively from one class have $100\%$ purity, and we refer to them as `objects' as opposed to the `snapshots'. 
Thus the Contrast-ClusterNet trained with the object-based approach is referred as ObjectNet, and is mainly for comparison purposes, even though it may have its own value if obtaining objects is a possibility.

\subsection{Self-supervised Feature Learning}
As we discussed in our previous work \citep{visapp21}, to verify the self-supervised feature learning of the SnapshotNet, we conduct experiments on both single-FOV snapshots and labeled objects derived from the scene - ‘Untermaederbrunnen3’ of the Semantic3D dataset.
The evaluations are based on the classification accuracy on the testing samples of an SVM (with a linear kernel) trained on the extracted features of training samples.
For the experiments, we train both the DGCNN and the SnapshotNet exclusively on the snapshot samples while keeping the ObjectNet trained on labeled objects.
For the SnapshotNet, we also want to see if the features can be applied to object samples. Thus we use the trained model to extract features of both snapshots and objects separately to train a different classifier, and this is referred to as ''SnapshotNet on snapshots" and ''SnapshotNet on objects" in Table \ref{scene1}.

\textbf{\textit{Learn with noises:}}
Table \ref{scene1} shows that, using 100\% of the training data, the DGCNN has the lowest accuracy compared to the other methods.
In comparison, the SnapshotNet tested on labeled objects has best performance on the total accuracy and all per-class accuracies except for low vegetation, on which is best performed by the SnapshotNet tested on snapshots.
Both the SnapshotNet and the ObjectNet yield a total accuracy above 97\%, and they are 10\% higher than the DGCNN.
This validates our claim that the proposed SnapshotNet is able to learn effective features from the raw point cloud complex scene in a self-supervised manner.
It also shows that the the noisy snapshots produce more powerful features by self-supervised learning than the fully-supervised DGCNN.

We see that the ObjectNet is able to achieve decent performance given that the training samples are derived from labeled data and each object sample has 100\% purity.
However, the snapshots used to train the SnapshotNet are often noisy.
As presented in Table \ref{semantic3d_stats}, some classes have low purity to start with, such as the Natural Terrain at a 57.47\% purity, the Scanning Artefacts at 73.2\% and the class of Cars at an 87.85\% purity.
While the snapshots from these classes are noisy, the SnapshotNet has shown high resistance over such noises in the data. (Note that there are no testing snapshots in natural terrain and scanning artefacts being sampled due to the small amount of data in the two classes)
The SnapshotNet on objects performs better than the ObjectNet, which is trained on noiseless objects.
It has also shown high accuracies over the aforementioned three noisy classes, which further confirms our hypothesis that powerful semantic features can be learned by predicting whether two segments are from the same snapshot and predicting the refined pseudo-labels for the snapshots, regardless of their semantic labels.  

\textbf{\textit{Classification with fewer labeled data:}}
To verify the effectiveness of the proposed weakly supervised classification, we gradually reduce the amount of labels involved in the training of the SVM.
The experiments are set up in the similar way by comparing the DGCNN, ObjectNet, and the SnapshotNet.

Table \ref{scene1} also shows how the classification accuracy of different models vary when the percentages of the classifier training data reduce from 100\% to 20\% and to 5\%.
It can be seen that the SnapshotNet outperforms the other models on overall accuracy.
The difference between the SnapshotNet (test on snapshots) and the DGCNN becomes more significant when the training data reduces, which spans from 11\% to 22.12\%.
Comparatively, the end-to-end fully-supervised DGCNN is very sensitive to the amount of training data due to the data hungry problem, that it needs sufficient labeled data to learn representative features.
The SnapshotNet however, doesn't rely on any labeled data for feature learning, and only need a small fraction of labeled data to train a classifier, which is a huge advantage over the fully-supervised model.

The snapshot-based SnapshotNet has also shown higher resistance on the reduction of the training data than the other training schemes: the accuracy only drops by 2.5\%, to 95\%, when the classifier is trained with merely 5\% of the data, while the object based SnapshotNet suffers a drop of 8.5\%, to 90.13\%, and the ObjectNet shows a bigger decline of 9.5\%, to 88.38\%.
We believe that the performance difference between the SnapshotNet and ObjectNet can be attributed to the use of noisy snapshots to make feature learning more robust.
Compared to the high purity objects, the snapshots forces the model to distinguish whether two segments are from the same area despite they might contain points from different classes.
In other words, we increase the difficulty of this pretext task by introducing noises and potentially leads to more representative features.

\begin{table*}[h!]
\centering
\footnotesize
\begin{tabular}{l|cc|cccccc}
\hline
\multicolumn{1}{c|}{\multirow{2}{*}{Method}} &
  \multirow{2}{*}{\begin{tabular}[c]{@{}c@{}}Overall\\ Accuracy (\%)\end{tabular}} &
  \multirow{2}{*}{\begin{tabular}[c]{@{}c@{}}Average\\ F-score (\%)\end{tabular}} &
  \multicolumn{6}{c}{Per-class F-scores (\%)} \\ \cline{4-9} 
\multicolumn{1}{c|}{} &               &               & Terrain       & Vegetation    & Building      & Hardscape     & Artefacts     & Cars          \\ \hline
Seg-aided\citep{guinard2017weakly}             & 83.3          & 82.3 & 98.1 & 67.0          & \textbf{98.8} & 91.5          & 51.3 & 82.3 \\
Pseudo-labelling\citep{pseudo_labelling}      & 95.6          & 66.7          & 94.2          & 61.2          & 97.7          & 84.6          & 9.0           & 53.3          \\ \hline
Part contrasting    & 96.9 & 80.2          & 97.4          & 85.5 & 98.4          & 95.1 & 32.3          & 72.5          \\
Scale contrasting      & 92.1          & 74.5          & 90.6          & 75.7          & 97.5          & 87.6          & 33.4          & 62.1          \\
Multi-FOV contrasting      & \textbf{97.6}          & \textbf{90.2}          & \textbf{98.2}          & \textbf{85.6}          & \textbf{98.8}          & \textbf{96.6}          & \textbf{74.8}          & \textbf{87.0}          \\ \hline
\end{tabular}
\caption{Semantic segmentation results on Semantic3D. Three self-supervised methods are compared against the state of the art weakly-supervised methods. All 8000 labels are used in the training of the classifier.}
\label{segmentation}
\end{table*}

\subsection{\textcolor{black}{Semantic Segmentation}}
Following the works of the seg-aided\citep{guinard2017weakly}, and pseudo-labelling\citep{pseudo_labelling}, and for the purpose of comparison, we merge the natural terrain and man-made terrain into one class of terrain, and put together the high vegetation and low vegetation as vegetation. 
\textcolor{black}{The experiments are carried out on these six classes cleaned from the Semantic3D dataset and the Oakland dataset, and they are organized as following:}

\begin{enumerate}

\item Having verified that the single-FOV snapshots, despite being noisy, are able to produce meaningful features from part-contrasting, we move forward to utilize the multi-FOV snapshot to feed the other two pre-text tasks: scale contrasting, and the multi-FOV contrasting, and comparison results are given.
\item For the weakly-supervised classifier, we experiment on three different ratios of the labeled data: 100\%, 20\%, 10\% and 5\% of our total 8000 labeled training samples, which contains 8000, 800, and 400 labels respectively.
These numbers of labels follow the distribution of the semantic labels in the dataset, therefore the larger classes might outnumber the small class on label numbers.
\item In addition to this, we include another test case with 30 labels per class, which is the same setup in the work of the seg-aided classification\citep{guinard2017weakly} and pseudo-labeling\citep{pseudo_labelling}. To mitigate the potential issues when using only a few labels, tests are conducted on the proposed cluster-based pseudo labeling for automatically adding more training samples. 
\item To evaluate the robustness of our method over different scene point clouds, the testing results are produced from two experimental setups.
One is to segment the scene point cloud from which the training samples are captured, while the other setup involves training the model on one scene but segmenting another one.
In the second case, we gradually add up the fine-tuning samples from the to-be-segmented scene to find a sweet spot where minimal fine-tuning is required to achieve comparable results when performing cross-scene semantic segmentation.

\end{enumerate}

\begin{figure*}[t!]%
\centering
\subfloat[Progress: 25\%; Iteration: 1820; F-score: 89.5\%]{\label{unter_groundtruth}\includegraphics[width=0.58\textwidth]{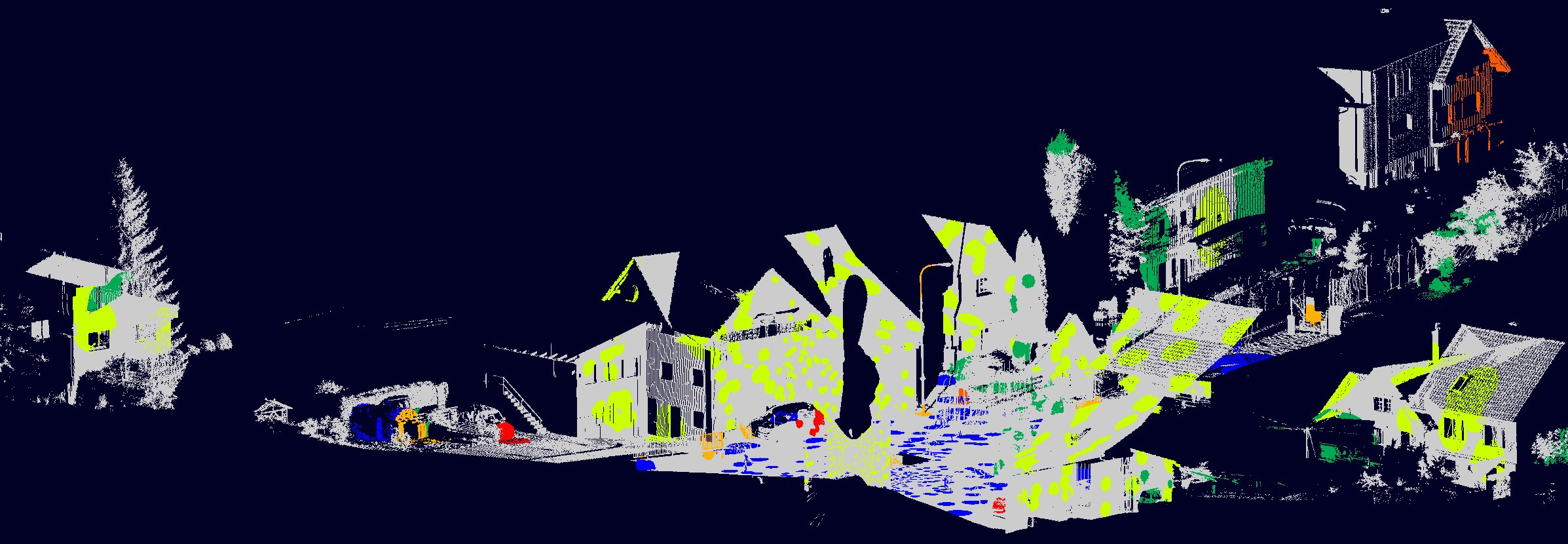}}

\centering
\subfloat[Progress: 50\%; Iteration: 4371; F-Score: 87.1\%]{\label{unter_groundtruth}\includegraphics[width=0.58\textwidth]{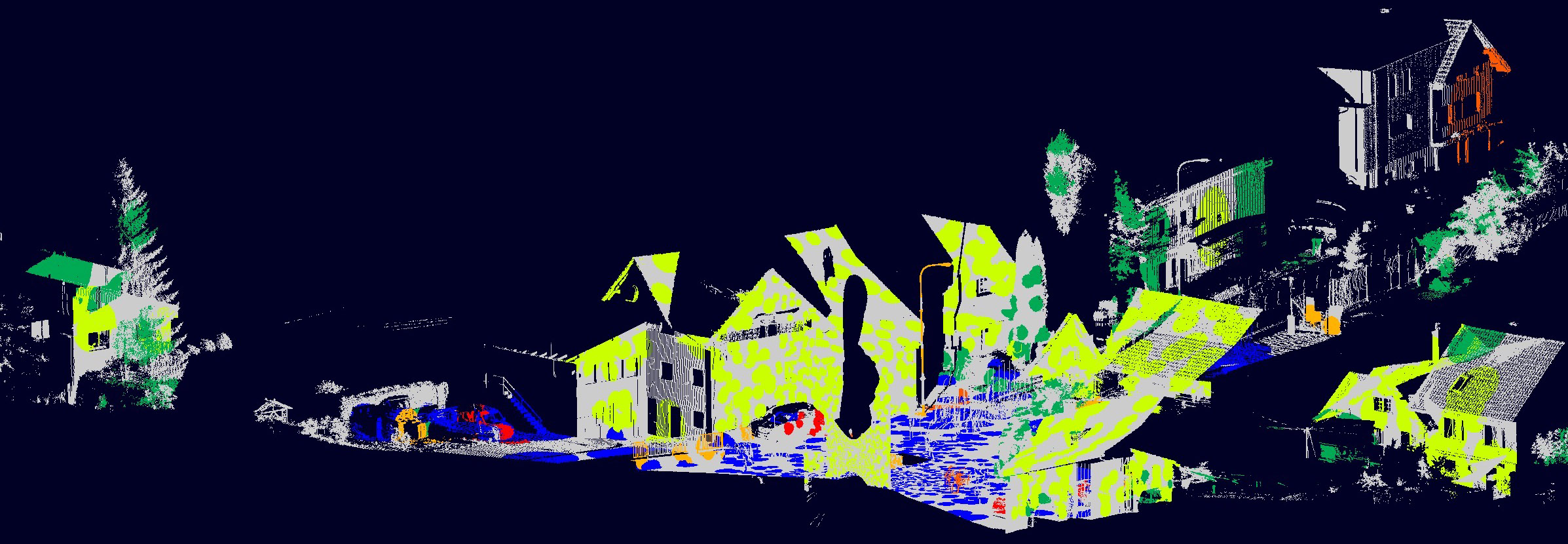}}

\centering
\subfloat[Progress: 75\%; Iteration: 8869; F-Score: 87.6\%]{\label{unter_groundtruth}\includegraphics[width=0.58\textwidth]{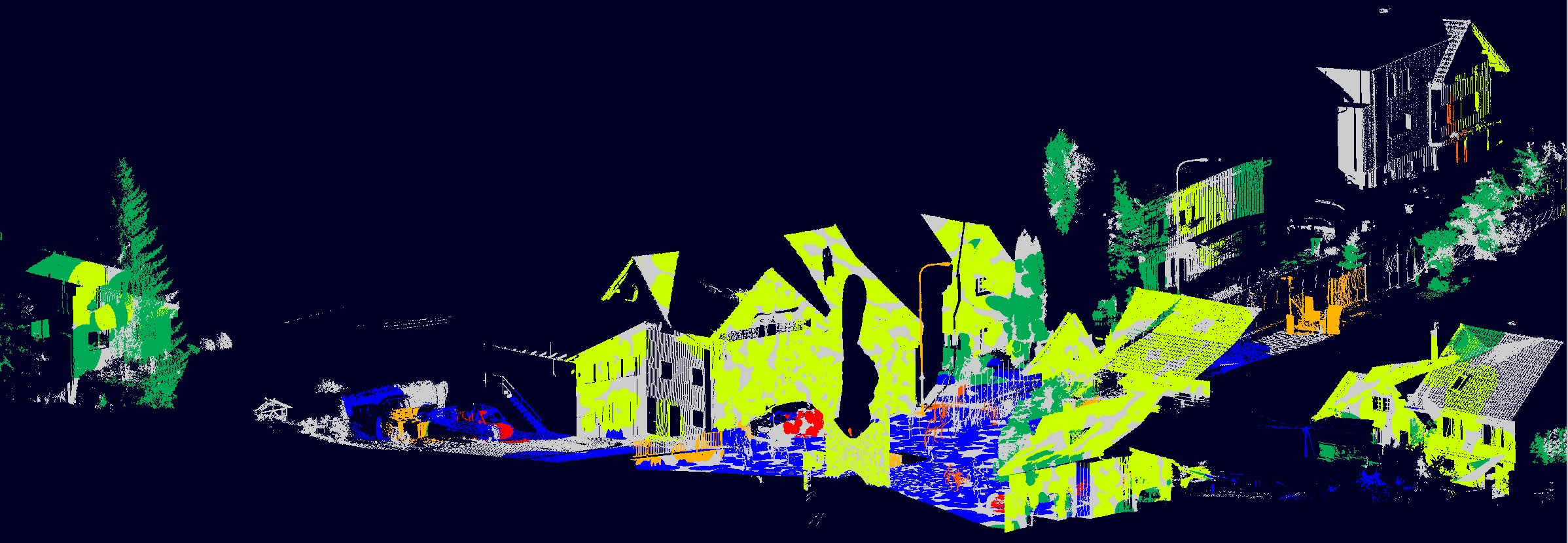}}

\centering
\subfloat[Progress: 99.95\%; Iteration: 84890; F-Score: 90.2\%]{\label{unter_groundtruth}\includegraphics[width=0.58\textwidth]{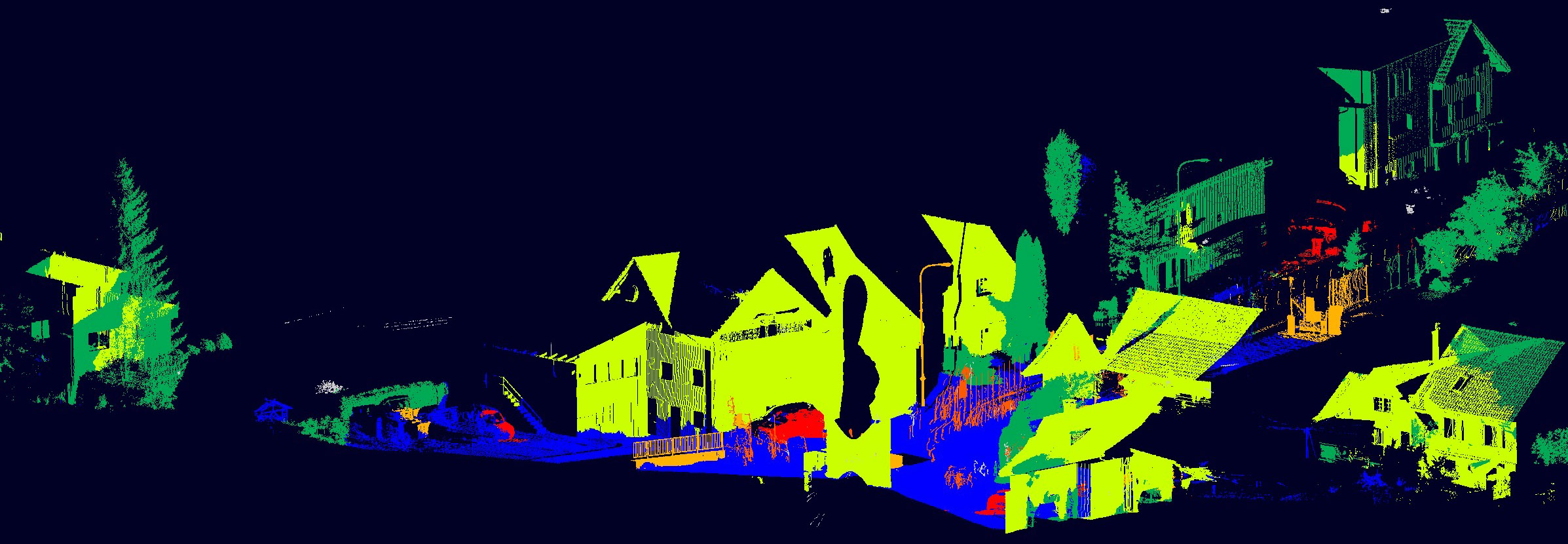}}
\centering

\subfloat[Ground truth]{\label{unter_groundtruth}\includegraphics[width=0.58\textwidth]{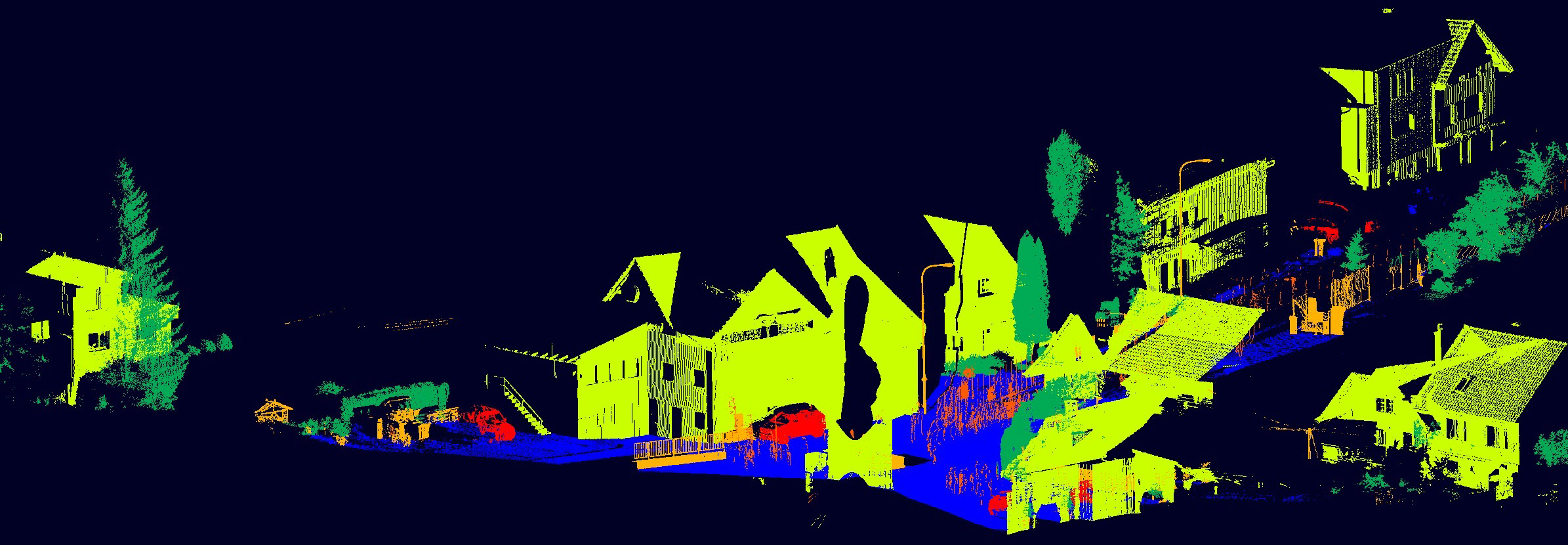}}
\caption{\textcolor{black}{Visualization of the progression of segmenting `Untermaederbrunnen3'. The model is governed with the pre-text of multi-FOV contrasting and the classifier is trained with 8000 labels. The bottom picture in \ref{unter_groundtruth} shows the ground truth to compare with. Colors correspond to semantic classes as the following: terrain is cyan, vegetation is green, buildings are yellow, hardscapes are orange, scanning artefacts are orange red, cars are red. The unlabeled points are in grey. We can see that snapshots gradually covers up the whole scene, during which previously falsely labeled points can be corrected by voting.}}

\end{figure*}

\begin{table*}[h!]
\centering
\footnotesize
\begin{tabular}{l|cc|cccccc}
\hline
\multicolumn{1}{c|}{\multirow{2}{*}{Method}} &
  \multirow{2}{*}{\begin{tabular}[c]{@{}c@{}}Overall\\ Accuracy (\%)\end{tabular}} &
  \multirow{2}{*}{\begin{tabular}[c]{@{}c@{}}Average\\ F-score (\%)\end{tabular}} &
  \multicolumn{6}{c}{Per-class F-scores (\%)} \\ \cline{4-9} 
\multicolumn{1}{c|}{} &
   &
   &
  Terrain &
  Vegetation &
  Building &
  Hardscape &
  Artefacts &
  Cars \\ \hline
Seg-aided\citep{guinard2017weakly} &
  83.3 &
  82.3 &
  98.1 &
  67.0 &
  98.8 &
  91.5 &
  51.3 &
  82.3 \\
Pseudo-labelling\citep{pseudo_labelling} &
  95.6 &
  66.7 &
  94.2 &
  61.2 &
  97.7 &
  84.6 &
  9.0 &
  53.3 \\ \hline
Ours 8000 labels &
  \textbf{97.6} &
  \textbf{90.2} &
  \textbf{98.2} &
  \textbf{85.6} &
  98.8 &
  \textbf{96.6} &
  \textbf{74.8} &
  \textbf{87.0} \\
Ours 1600 labels &
  96.9 &
  84.9 &
  97.3 &
  84.0 &
  \textbf{98.9} &
  95.1 &
  57.5 &
  76.9 \\
Ours 800 labels &
  95.9 &
  82.6 &
  96.3 &
  79.7 &
  98.3 &
  93.6 &
  59.8 &
  67.7 \\
Ours 400 labels &
  94.2 &
  73.3 &
  94.6 &
  76.0 &
  97.9 &
  92.6 &
  31.4 &
  47.8 \\
Ours 180 labels &
  85.4 &
  63.4 &
  84.0 &
  57.5 &
  94.3 &
  89.2 &
  11.5 &
  44.3 \\\hline
\end{tabular}
\caption{Parameter studies on varying the numbers of labels involved in training. Experiments are performed on our method with the multi-FOV contrasting to compare against the state of the art methods on Semantic3D. The number of labels corresponds to 100\%, 20\%, 10\%, and 5\% of the total available labels. An additional test case using 30 labels/class is included to make a total 180 labels.}
\label{weakly_segmentation}
\end{table*}

\begin{table*}[h!]
\centering
\footnotesize
\begin{tabular}{l|c|ccccc}
\hline
\multicolumn{1}{c|}{\multirow{2}{*}{Method}} &
  \multirow{2}{*}{\begin{tabular}[c]{@{}c@{}}Overall\\  Accuracy (\%)\end{tabular}} &
  \multicolumn{5}{c}{Per-class F-scores (\%)} \\ \cline{3-7} 
\multicolumn{1}{c|}{}                       &      & Scatter Misc & Default Wire & Utility Pole & Load Bearing & Facade \\ \hline
Seg-aided\citep{guinard2017weakly}        & 96.6 & 93.7         & 46.5         & 8.7          & 99.5         & 93.9   \\
Pseudo-labelling\citep{pseudo_labelling} & 96.6 & 92.0         & 40.2         & 46.2         & 99.3         & 93.3   \\ \hline
Ours 8000 labels                            & 92.6 & 85.7         & 2.5          & -            & 96.7         & 84.4   \\
Ours 800 labels                             & 92.2 & 84.6         & 0.2          & -            & 96.7         & 83.2   \\
Ours 400 labels                             & 91.3 & 82.6         & 6.9          & -            & 96.1         & 82.2   \\
Ours 200 labels                             & 89.4 & 78.6         & -            & -            & 95.7         & 73.8   \\
Ours 150 labels                             & 88.5 & 76.0         & -            & -            & 95.1         & 68.4   \\ \hline
\end{tabular}
\caption{\textcolor{black}{Parameter studies on 3D scene segmentation with varying the numbers of labels involved in training on the Oakland outdoor dataset. Experiments are performed on our method with the multi-FOV contrasting to compare against the state of the art methods on the Oakland dataset. The number of labels corresponds to 100\%, 10\%, 5\%, and 2.5\% of the total available labels. An additional test case is included using a total of 150 labels randomly drew from the 8000 training samples. Note that ``-'' suggests no training samples are available for a particular class.}}
\label{oakland_weakly_segmentation}
\end{table*}

\textbf{\textit{Results on various contrastive learning:}}
The results of the three contrasting approaches are listed in Table \ref{segmentation}, where our method with the three approaches is compared with a state of the art method seg-aided classification \citep{guinard2017weakly} and one comparable following study, the pseudo-labeling approach\citep{pseudo_labelling}.
By doing part contrasting, our model yields an overall accuracy (OA) at 96.9\%, 1.3\% higher than the pseudo-labeling method.
The average F-score, however, is slightly lower than the Seg-aided classification at 80.2\%.
Looking at the per-class F-scores, the part contrasting produces comparable results with the seg-aided method on the classes of terrain and building, ours shows a prominent improvement over the vegetation and hardscape classes, which pushes up the per-class F-score by 18.5\% and 3.6\% respectively.
However, the part contrasting performs noticeably worse than the seg-aided on small objects such as artefacts and cars.
This seems to conform our conjecture that features learned by part contrasting are vulnerable when describing smaller items.

The scale contrasting is proposed to bring the features at a more abstract level into the play, thus to achieve better results on the small objects in the scene.
There is a marginal increase on the F-score of the artefacts than the part contrasting, while at the cost of worsening on all other classes.
Another small class of cars is also 10.4\% worse than the part contrasting and 20.2\% lower than the seg-aided method.
So far the results have been suggesting that the scale contrasting might lack the capability of pushing hard for powerful versatile features, and the produced features are less descriptive at certain levels that are vital to to distinguish larger objects.

With this finding, the scale contrasting and part contrasting are combined for further experiments, seeking to strengthening the features from both perspectives.
The collaborative effort of the part contrasting and scale contrasting urges the model to develop powerful features leveraging knowledge from both the object level information and higher level structural information.
Our method equipped with the multi-FOV contrasting outperforms the pseudo-labeling method on the OA by 2\% at 97.6\%, and on F-score by 23.5\%; it is 14.3\% above the Seg-aided classification\citep{guinard2017weakly} on OA, and a gain of the F-score is seen at 7.9\%.
There are some significant improvement over the small classes: the per-class F-score of the artefacts is 23.5\% higher than the state of the art method at 74.8\%.
The class of cars has seen an increase from 82.3\% to 87\% using our method.
The classes of vegetation and hardscape also experienced a very noticeable boost on their F-scores, and the terrain and building have a slight edge over the state of the art performance by the Seg-aided classification.

\textbf{\textit{\textcolor{black}{Segmentation with fewer labels on Semantic3D dataset:}}}
Table \ref{weakly_segmentation} shows the effect of reducing the numbers of available labels during the training of the SVM classifier on the Semantic3D.
Our method with the multi-FOV contrasting is tested against the seg-aid classification and the pseudo-labeling, both using 30 labels per class.
When the training data is reduced ten times to 800 labeled samples, similar performances can still be observed despite the drastic drop in the available training data.
The OA of the 800 labels model is marginally greater than the pseudo-labeling approach by 0.3\%, the average F-score is 15.9\% higher than the pseudo-labeling approach and 0.3\% over the best result from the seg-aided classification.
The per-class F-score is still significantly greater than the state of the art method on the vegetation and artefacts, and is in the lead on the class of hardscape.
The cars, however is 14.6\% lower than the segmentation-aided classification at this level of available labels.
A steady growth on this class can still be seen when the labels increase and we expect a surpassing over the seg-aided method when it is trained with more than 1600 labeled data.
When further reducing the labeled data by half, the overall performance starts to drop.
The per-class F-score on artefacts experiences a 28.4\% decrease and for the cars it also falls by 19.9\%.
This suggests that a further cut down on labeled data usage by only 400 might come at a high price.

While our method has shown a superiority over the SOA methods when using as few as 800 labels, the SOA methods are only tested on 30 labels per class, making it 180 labels in total.
To make a fair comparison with the two baseline methods, the training data is reduced to 30 labels per class.
Apart from the artefacts holding a slight advantage over the pseudo-labeling method by 2.5\%, the other classes are below the two baselines by varying degrees.
Further investigating into the cause of this significant deficiency, two explanations are speculated.
As commonly occurred in machine learning, underfitting of the classifier might be a big factor on the poor performance.
A solution to this issue is to deploy larger amount training data.
Another factor here is that, our self-supervised learning is trained on only 8000 samples from a particular statistical distribution in terms of their semantic labels, meaning that the learned features might not be as equally weighted, and the weights seem to collapse when training the classifier with uniformly distributed labels.
Looking into more details, the recall rate of the artefacts is 86.0\% while the precision is merely 6.2\%.
On the other hand, the recall of the terrain is 75.7\% but the precision remains as high as 94.2\%.
These extremes are also seen on the vegetation and cars.
This seems to meet our conjecture that the even labels bring bias into some of these classes.

\textbf{\textit{\textcolor{black}{Segmentation with fewer labels on Oakland dataset:}}}
\textcolor{black}{
Similar experiments are further conducted on the Oakland dataset. 
The results in Table \ref{oakland_weakly_segmentation} illustrates the OA and per class F-score against the two baselines.
The Average F-score is not included in this test due to the lack of training samples for the disadvantaged classes such as default wire and utility pole.
The Oakland dataset is extremely unbalanced between different classes and among the 8000 samples there are only 10 default wire snapshots and 6 utility pole snapshots.
So when randomly selecting a smaller group of training samples from the pool, these classes become absent during the training, and these corresponding per-class F-scores are noted as ``-''.}

\textcolor{black}{
With the available results, we can still exploit some insights towards our model's characteristics.
The overall accuracy gradually decreases when using fewer labels, however the scatter misc and facade classes experience a steeper decline than the load bearing class does.
This trend conforms the observation from the Semantic3D experiments, that the dominant classes have higher resistance on performance decay over the disadvantaged classes.
Unlike the Semantic3D experiments, our model does not surpass the SOA methods with all 8000 samples.
Referring to our previous work on this problem\citep{visapp21}, when training with single-FOV snapshots, the classification accuracy on the Oakland dataset using all training samples is 90.6\% and it is 96.4\% on Semantic3D.
This confirms that the segmentation results are consistent with the classification accuracy, despite being benefited from the new multi-FOV training.
So the performance difference between the two experiment setups reflects our model's reaction towards these two datasets.}

\textcolor{black}{
Here some interpretations are made to account for this performance discrepancy: 
Both experiments have shown the advantage of our model on feature learning with the use of snapshots.
However, the way we generate these snapshots is fairly straightforward and fully statistical by randomly taking k nearest neighbors.
This poses a problem when some classes are loosely distributed and outnumbered by other classes in the background, such as the default wire and utility pole in this scenario.
These two classes are barely sampled as a snapshot by our definition of ``snapshots'', and therefore exacerbates the inequality of the training samples than it appears in the entire scene.
}

\begin{table}[]
\small
\begin{tabular}{|c|c|c|}
\hline
Threshold & Number of Pseudo Labels & Accuracy in Labeling \\ \hline
0.9       & 2631                    & 98.1\%               \\ \hline
0.8       & 4253                    & 96.1\%               \\ \hline
0.75      & 5078                    & 93.7\%              \\ \hline
\end{tabular}
\caption{\textcolor{black}{Trade-off of the threshold section in the Cluster-based pseudo labeling on Semantic3D. 120 clusters are randomly selected from the total 300 clusters for pseudo labeling.} Larger threshold value puts heavier constraints on the pseudo-labeling, leading to fewer labels but higher labeling accuracy.}
\label{semantic3d_threshold}
\end{table}

\begin{table*}[h!]
\centering
\footnotesize
\begin{tabular}{l|cc|cccccc}
\hline
\multicolumn{1}{c|}{\multirow{2}{*}{Method}} &
  \multirow{2}{*}{\begin{tabular}[c]{@{}c@{}}Overall\\ Accuracy (\%)\end{tabular}} &
  \multirow{2}{*}{\begin{tabular}[c]{@{}c@{}}Average\\ F-score (\%)\end{tabular}} &
  \multicolumn{6}{c}{Per-class F-scores (\%)} \\ \cline{4-9} 
\multicolumn{1}{c|}{} &
   &
   &
  Terrain &
  Vegetation &
  Building &
  Hardscape &
  Artefacts &
  Cars \\ \hline
Seg-aided\citep{guinard2017weakly} &
  83.3 &
  82.3 &
  98.1 &
  67.0 &
  98.8 &
  91.5 &
  51.3 &
  82.3 \\
Pseudo-labelling\citep{pseudo_labelling} &
  95.6 &
  66.7 &
  94.2 &
  61.2 &
  97.7 &
  84.6 &
  9.0 &
  53.3 \\ \hline
180 labels without pseudo labeling &
  85.4 &
  63.4 &
  84.0 &
  57.5 &
  94.3 &
  89.2 &
  11.5 &
  44.3 \\
120 clusters t0.9 + 10 labels/class &
  91.6 &
  70.6 &
  91.6 &
  61.8 &
  96.9 &
  87.2 &
  32.6 &
  53.4 \\
120 clusters t0.8 + 10 labels/class &
  \textbf{92.2} &
  \textbf{74.1} &
  91.6 &
  71.8 &
  96.9 &
  87.4 &
  32.5 &
  64.7 \\
120 clusters t0.75 + 10 labels/class &
  92.0 &
  73.5 &
  91.1 &
  67.3 &
  97.3 &
  85.3 &
  31.1 &
  69.1 \\\hline
\end{tabular}
\caption{Parameter studies on the cluster-based pseudo labeling threshold selection on Semantic3D. Experiments are performed on our method with the multi-FOV contrasting to compare against the state of the art methods. Three threshold levels are tested on 120 random clusters for pseudo labeling samples. An test case using 30 labels/class without pseudo labeling is included for comparison.}
\label{semantic3d_cluster_based_pseudo}
\end{table*}

\begin{table*}[h!]
\centering
\footnotesize
\begin{tabular}{l|cc|c|ccccc}
\hline
\multicolumn{1}{c|}{\multirow{2}{*}{Method}} &
  \multicolumn{1}{l}{\multirow{2}{*}{\begin{tabular}[c]{@{}l@{}}Number of \\ Pseudo Labels\end{tabular}}} &
  \multicolumn{1}{l|}{\multirow{2}{*}{\begin{tabular}[c]{@{}l@{}}Labeling \\ Accuracy(\%)\end{tabular}}} &
  \multirow{2}{*}{\begin{tabular}[c]{@{}c@{}}Overall\\ Accuracy (\%)\end{tabular}} &
  \multicolumn{5}{c}{Per-class F-scores (\%)} \\ \cline{5-9} 
\multicolumn{1}{c|}{} & \multicolumn{1}{l}{} & \multicolumn{1}{l|}{} &               & Scatter Misc  & Default Wire & Utility Pole & Load Bearing  & Facade        \\ \hline
150 random labels     & 0                    & 100                   & 88.5          & 76.0          & -            & -            & 95.1          & 68.4          \\ \hline
150 clusters t0.99    & 312                  & 97.4                  & 78.2          & 59.4          & -            & -            & 87.7          & 59.4          \\
150 clusters t0.97    & 762                  & 90.6                  & \textbf{81.4} & \textbf{65.5} & -            & -            & \textbf{89.0} & \textbf{66.6} \\
150 clusters t0.95    & 1479                 & 89.3                  & 78.7          & 58.5          & -            & -            & 88.4          & 64.4          \\
150 clusters t0.90    & 3121                 & 87.7                  & 76.8          & 56.1          & -            & -            & 86.0          & 64.6          \\ \hline
\end{tabular}
\caption{\textcolor{black}{Parameter studies on the cluster-based pseudo labeling threshold selection and segmentation results on the Oakland dataset. Four threshold levels are tested on 150 random clusters for pseudo labeling samples. An additional test case using 150 random labels without pseudo labeling is included for comparison.}}
\label{oakland_cluster_based_pseudo}
\end{table*}

\textbf{\textit{Results on cluster-based pseudo labeling:}}
To mitigate the impacts of the aforementioned two potential issues when using only a few labels, tests are conducted on the proposed cluster-based pseudo labeling for more training samples.
Here on the Semantic3D, 120 clusters are randomly selected out of 300 and each cluster center is assigned with one label.
In addition, a collection of 10 labeled sampled from each class is joined into the pseudo labeled data, consuming a total of 180 labels.
Table \ref{semantic3d_threshold} illustrates a trade-off effect between the number of pseudo labels and the labeling accuracy from the choice of threshold.
A larger threshold causes a heavier constraint when selecting the samples near the cluster center, which leads to fewer samples to be pseudo-labeled but they are essentially much more likely to share the same semantic label with the cluster center.
According to table \ref{semantic3d_cluster_based_pseudo}, comparing to our method using 180 labels without pseudo labeling, a boost of 7\% on the overall accuracy and 11\% on the F-score is seen, when thresholding at 0.8. 
Our cluster-based pseudo labeling method has outperformed the SOA deep learning method using pseudo-labelling on the F-score by 7.4\%, particularly ours has an edge on segmenting the smaller objects such as artefacts and cars, where increases of 23.5\% and 11.4\% are gained respectively.
When comparing to the seg-aided method, we also have an advantage on the overall accuracy by 8.9\%.

It is worth explaining on the decision of the random selection when choosing 120 clusters for pseudo labeling.
It was realized that for those larger clusters, despite being able to generate more pseudo labels, the included samples tend to be homogeneous. 
In other words, their features are less representative, so it’s not always a good idea to go for larger clusters in the pursuit of more pseudo labels. 
For instance, two buildings looking completely different might have their samples far from each other in the feature space, and it would be more helpful to have each of their samples being chosen during the pseudo labeling.
On the other hand, the very small clusters often contain few samples from the minority classes, and these clusters are an important source of acquiring distinct features from those minority classes.
These factors pose a difficult decision on the cluster selection, to obtain as many pseudo labels as possible while maintain the diversity of the features.
Thanks to the large number of clusters, we believe that random selection is a better way to evenly include both large and small clusters for pseudo labeling.

\textcolor{black}{
The results on the Oakland dataset provide a better viewpoint on the label number vs. labeling accuracy dilemma when the dataset is highly unbalanced.
Table \ref{oakland_cluster_based_pseudo} shows this trade-off and the leading segmentation performance using 150 labels.
The baseline is when randomly choosing 150 labeled samples without pseudo labeling, and it is worth noting that since each sample has three FOVs, so there are a total of 450 training snapshots in this case.
When loosing the threshold on the pseudo labeling, more pseudo labels are obtained at the cost of lower labeling accuracy.
At 97\%, the segmentation performance reaches the best among all four pseudo labeling tests, where we have 762 pseudo labels at the accuracy of 90.6\%.
By comparing this result with the baseline, it is seen that 1.7 times more training samples cannot makeup for the 9.4\% loss on labeling accuracy, and it is not a problem on the Semantic3D where thousands of pseudo labels are obtained with high accuracy.
}

\begin{figure*}[t!]
    \subfloat[Progress: 99.95\%; F-score: 84.7\%]{
    \begin{minipage}[c]{0.48\textwidth}
       \centering
       \includegraphics[width=0.98\textwidth]{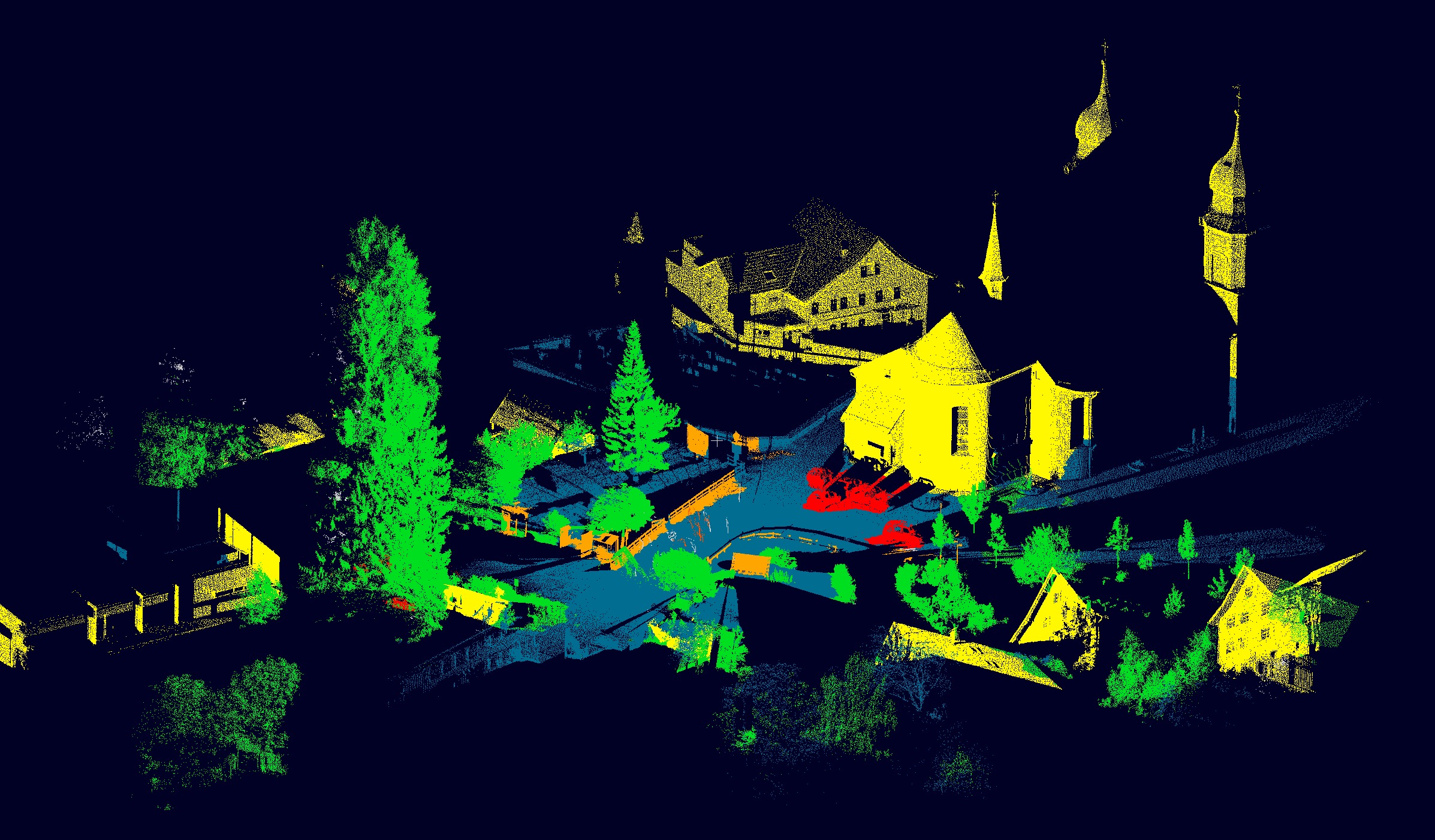}
    \end{minipage}}
    \subfloat[Ground truth]{
    \begin{minipage}[c]{0.48\textwidth}
       \centering
       \includegraphics[width=0.98\textwidth]{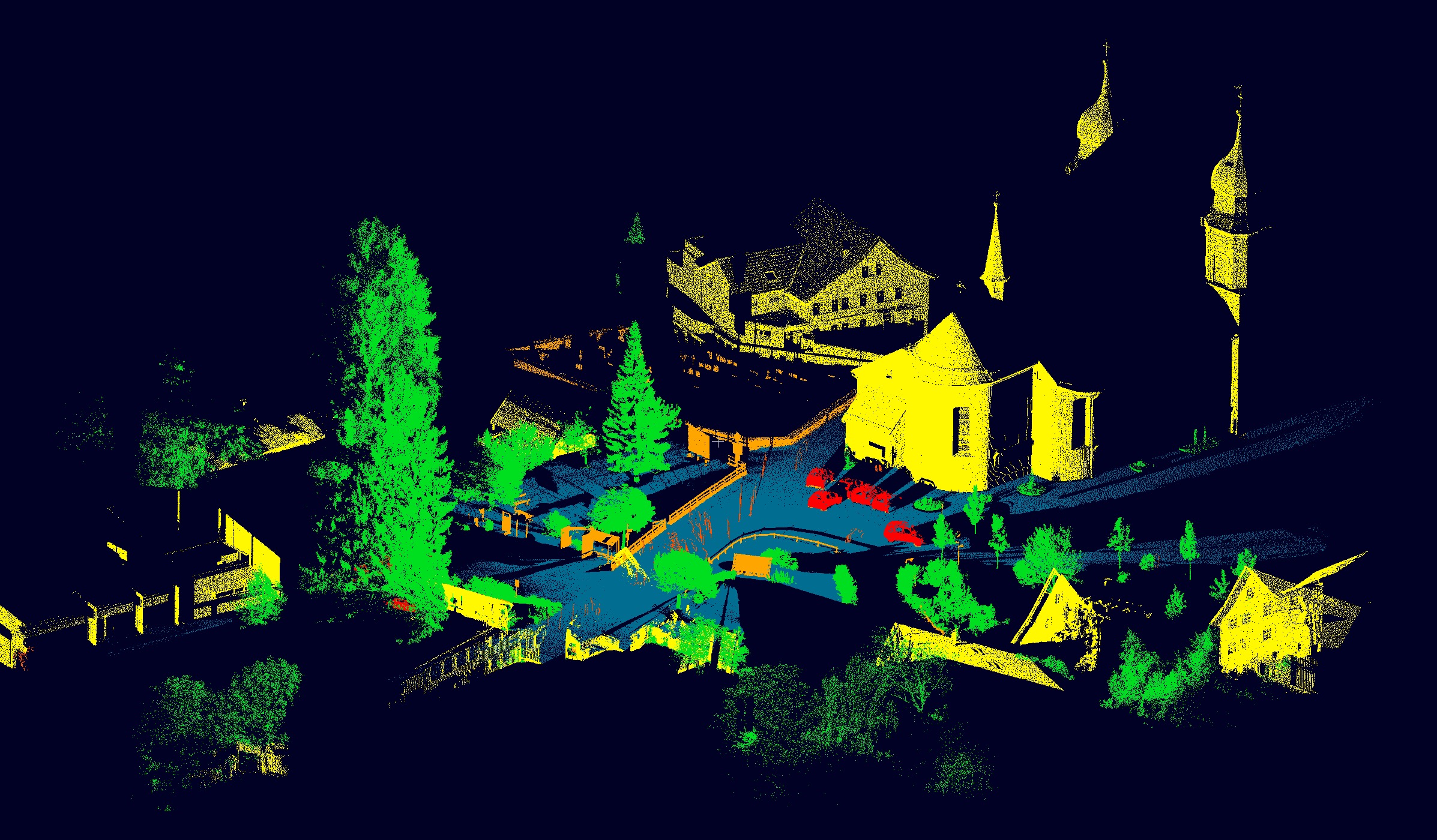}
    \end{minipage}}
    \caption{Semantic labelling of the scene `Bildstein3' using the model trained on `Untermaederbrunnen3'. Colors correspond to semantic classes as the following: terrain is cyan, vegetation is green, buildings are yellow, hardscapes are orange, scanning artefacts are orange red, cars are red, and the unlabeled points are grey. It can be seen that our method is capable of recognizing the rough outline of the smaller items such as cars, but lacks precise semantic labeling.}
    \label{cross_scene_visual}
\end{figure*}

\textbf{\textit{Segmentation across scenes:}}
Having shown the advantages of proposed method on a single scene point cloud when the learning is governed by the multi-FOV contrasting.
To verify the strength of our model on quickly adapting to other data, the following experiments are designed to test on cross-scene segmentation.
The goal is to find out if this model is capable of producing decent segmentation performance by only  fine-tuning  the model on minimal amount of fine-tuning data.
Taking the model obtained from previous experiments, which is trained on the scene `Untermaederbrunnen3' from the Semantic3D.
This model is fine-tuned with a series of number of samples from the scene `Bildstein3', such as 1600, 800, 400, 0 samples, which take up to 20\%, 10\%, 5\%, and 0\% (no fine-tuning) of the total samples.
Recall that the model has two networks working in sequence.
The clusternet is trained with the pseudo-labels acquired from the features extracted using a well trained contrastnet.
So to fine-tune our model with new data, the pseudo-labeling process is carried through first.
This involves extracting features of the fine-tuning data using a pre-trained contrastnet and predicting new pseudo-labels for them with the converged KMeans from our previous experiments.
Then the pre-trained clusternet is fine-tuned on the new data with their pseudo-labels before starting segmenting the new scene.

\begin{table*}[]
\centering
\begin{tabular}{c|cc|cccccc}
\hline
\multirow{2}{*}{Fine-tune data} &
  \multirow{2}{*}{\begin{tabular}[c]{@{}c@{}}Overall\\ accuracy (\%)\end{tabular}} &
  \multirow{2}{*}{\begin{tabular}[c]{@{}c@{}}Average\\ F-score (\%)\end{tabular}} &
  \multicolumn{6}{c}{Per-class F-scores (\%)} \\ \cline{4-9} 
     &      &      & Terrain & Vegetation & Building & Hardscape & Artefacts & Cars \\ \hline
1600 & 93.2 & 84.7 & 94.3    & 94.9       & 94.7     & 87.8      & 47.8      & 88.7 \\
800  & 96.2 & 84.8 & 97.1    & 95.4       & 97.5     & 94.7      & 36.9      & 87.9 \\
400  & 94.5 & 85.5 & 95.6    & 95.1       & 94.3     & 90.7      & 47.4      & 89.6 \\
0    & 95.9 & 84.7 & 96.8    & 95.4       & 95.2     & 94.4      & 33.4      & 93.4 \\ \hline
\end{tabular}
\caption{Cross scene segmentation performance of our method. The model is pre-trained on the scene ‘Untermaederbrunnen3’ and fine-tuned on different numbers of samples from ‘Bildstein3’.The decreasing numbers of fine-tune data make up to 20\%, 10\%, 5\%, and 0\% of the total 8000 samples captured from ‘Bildstein3’.}
\label{finetune}
\end{table*}

As demonstrated in Table \ref{finetune}, experiments are tested with different amount of fine-tune data from 1600 to 0 samples.
It is seen that the OA and average F-scores are close to each other among the four fine-tuned models.
Compared to the other five classes, artefacts are prone to large fluctuations on F-score.
To compare with the single-scene performance trained with 8000 labels (Table \ref{weakly_segmentation}), a drop of 27\% on the artefacts attracts most attention among all other results, which are considerably close to or even surpasses the single-scene results.
Despite that the results have shown an edge of our method when adopting to new data even without any fine-tuning, there is no significant improvement by adding in more fine-tuning data.
One interpretation of this particular behavior is again related to the statistical distribution of the classes.
As discussed on the snapshot purity, the number of points in each class are greatly uneven, leading to a large disparity on the number of snapshots being captures in each class.
This is particularly the case when it comes to the artefacts, where as shown in Table \ref{fig:snapshots} that no snapshots from this class are picked up during sampling, meaning that the fine-tuning was largely conducted on samples from bigger classes such as terrain or building.
Figure \ref{cross_scene_visual} illustrates the visualization of the semantic labelling compared with the ground truth.
It can be observed that the hardscapes away from the center are mislabeled as terrain, and this also happens on the lower part of the church's tower.
The cars are correctly located but the labeling precision is not as satisfactory because the surrounding terrains are mislabeled into cars.
These observations again align with our hypothesis, that some classes are under-trained due to the lack of samples or the training is contaminated with low-quality samples.

\section{Conclusion and Discussion}
\label{sec:conclusion}
In summary, we have proposed the SnapshotNet for self-supervised feature learning on the complex scene point cloud, including a new pre-text task that joins the part contrasting and the proposed scale contrasting for stronger features.
We have also designed a weakly-supervised method for point cloud semantic segmentation by training with fewer coarse-grained labels.
While reducing the labels involved in the downstream tasks, a cluster-based pseudo labeling technique is implemented to obtain more training data.
\textcolor{black}{The proposed methods are evaluated and verified on two real life complex scene datasets} and the experimental results indicate that our method is capable of learning effective features from unlabeled scene point cloud data.
Compared to the state of the art methods, our methods still show several advantages.
This model is able to produce comparable results at a slightly higher cost on label collection.
When the cluster-based pseudo labeling is enabled on representative snapshots, our model is capable of producing comparable results with the state-of-the-art methods using only 180 labels.
As a deep learning model, our method does not rely on hand crafted features, and it has proven to be robust to be directly applied to cross-scene segmentation without or with a small dose of fine-tuning within the same dataset, saving the effort of training a model on every new scene.

There are also some weaknesses of the proposed method, particularly that the quality of snapshots capturing are greatly influenced by the statistical distribution of the semantic classes.
We have tried to resolve this issue by designing the multi-FOV snapshots and have gained some significant improvement, but the performance on smaller items still needs further improvement.
In the future, the snapshot capturing could be further investigated, such as utilizing the surface normal or other local geometrical information, to potentially improve the sampling quality and enhancing the local semantic labelling precision.
The noises in the snapshot sampling could also be turned into certain advantages for a multi-level contrastive learning: noisy snapshots might contain parts from other objects such that not all points from both samples agree on each other when forming a positive pair.
If the dissimilarity between two parts can be quantified and well measured, a contrastive pair can be then defined in one of the multiple levels, instead of choosing from only positive or negative.

\section{\uppercase{Acknowledgments}}
\label{sec:acknowlegments}
The research is supported by the National Science Foundation through Awards PFI \#1827505 and SCC-Planning \#1737533, the Air Force Office of Scientific Research (AFOSR) via Award \#FA9550-21-1-0082, and Bentley Systems, Incorporated, through a CUNY-Bentley CRA 2017-2020. Additional support is provided by a CCNY CEN Course Innovation Grant from Moxie Foundation, and the Intelligence Community Center of Academic Excellence (IC CAE) at Rutgers University (Awards \#HHM402-19-1-0003 and \#HHM402-18-1-0007).

\bibliographystyle{model2-names}
\bibliography{refs}

\end{document}